\documentclass[preprint,5p,twocolumn]{elsarticle}

\usepackage{amssymb}
\usepackage{amsmath}

\usepackage{algorithmic}
\usepackage{algorithm}
\usepackage{textcomp}
\usepackage{stfloats}
\usepackage{url}
\usepackage{verbatim}
\usepackage{graphicx}
\usepackage{ulem}
\usepackage{booktabs}
\usepackage{multirow}
\usepackage{hyperref}
\usepackage{pifont}
\usepackage{array}
\usepackage{threeparttable}
\usepackage[caption=false,font=footnotesize,labelfont=rm,textfont=rm]{subfig}

\journal{xxx}
\begin{document}

\begin{frontmatter}



\title{Mind the Gap: Towards Generalizable Autonomous Penetration Testing via Domain Randomization and Meta-Reinforcement Learning}


\author[inst1,inst2]{Shicheng Zhou }
            \ead{zhoushicheng@nudt.edu.cn}
		\author[inst1,inst2,inst3]{Jingju Liu \corref{cor1}}
            \ead{liujingju17@nudt.edu.cn}
            \author[inst1,inst2]{Yuliang Lu \corref{cor1}}
            \ead{luyuliang@nudt.edu.cn}
            \author[inst3]{Jiahai Yang}
            \ead{yang@cernet.edu.cn}
            \author[inst4]{Yue Zhang}
            \ead{zhangyue@nudt.edu.cn}
            \author[inst1]{Jie Chen}
            \ead{jchen202209@163.com}
		\affiliation[inst1]{organization={College of Electronic Engineering, National University of Defense Technology},
			city={Hefei},
			postcode={230037}, 
			country={China}}

            \affiliation[inst2]{organization={Anhui Province Key Laboratory of Cyberspace Security Situation Awareness and Evaluation},
			city={Hefei},
			postcode={230037}, 
			country={China}}
		\affiliation[inst3]{organization={Institute for Network Sciences and Cyberspace and the Beijing National Research Center for Information Science and Technology, Tsinghua University},
			city={Beijing},
			postcode={10084}, 
			country={China}}
		\affiliation[inst4]{organization={College of Computer Science and Technology, National University of Defense Technology},
			city={Changsha},
			postcode={410073}, 
			country={China}}
        \cortext[cor1]{Corresponding authors}
\begin{abstract}

With increasing numbers of vulnerabilities exposed on the internet, autonomous penetration testing (pentesting) has emerged as a promising research area. Reinforcement learning (RL) is a natural fit for studying this topic. 
However, two key challenges limit the applicability of RL-based autonomous pentesting in real-world scenarios: (a) training environment dilemma -- training agents in simulated environments is sample-efficient while ensuring their realism remains challenging; (b) poor generalization ability -- agents' policies often perform poorly when transferred to unseen scenarios, with even slight changes potentially causing significant generalization gap. 
To this end,  we propose GAP, a generalizable autonomous pentesting framework that aims to realizes efficient policy training in realistic environments and  train generalizable agents capable of drawing inferences about other cases from one instance.
GAP introduces a Real-to-Sim-to-Real pipeline that (a) enables end-to-end policy learning in unknown real environments while constructing realistic simulations; (b) improves agents' generalization ability by leveraging domain randomization and meta-RL learning.
Specially, we are among the first to apply domain randomization in autonomous pentesting and propose a large language model-powered domain randomization method for synthetic environment generation. We further apply meta-RL to improve agents' generalization ability in unseen environments by leveraging synthetic environments.  The combination of two methods effectively bridges the generalization gap and improves agents' policy adaptation performance.
Experiments are conducted on various vulnerable virtual machines, with results showing that GAP can enable policy learning in various realistic environments,  achieve zero-shot policy transfer in similar environments, and realize rapid policy adaptation in dissimilar environments.  The source code is available at \url{https://github.com/Joe-zsc/GAP}.  

\end{abstract}




\end{frontmatter}




\section{Introduction}
Penetration testing,
shortly pentesting or PT, is an effective methodology to assess cybersecurity through authorized simulated cyber attacks.  It aims to preemptively identify security vulnerabilities, allowing organizations to proactively enhance their security measures and defenses.  
However, with more and more vulnerabilities exposed on the internet, traditional manual-based pentesting has become more costly, time-consuming, and personnel-constrained \cite{Lore}.  In light of this, autonomous pentesting has emerged as a promising research area.
Pentesting is a dynamic sequential decision-making process, while reinforcement learning (RL) is a suitable method for optimizing such decisions. RL trains an agent to learn a policy through trial and error by interacting with the environment, without the need for supervision or predefined environmental models.  This makes RL a natural fit for studying autonomous pentesting.

RL-based autonomous pentesting aims to train agents to learn how to explore and exploit vulnerabilities in target hosts. 
However, achieving satisfactory performance with  RL algorithms often necessitates a large number of training samples \cite{ChenHJLW22}. This is typically impractical in autonomous pentesting, where agents interacting with real environments are time-consuming and risky due to action execution and network latency. 
A common way to tackle this issue is to train the agent in simulated or emulated environments and subsequently transfer the learned policy to real-world scenarios. But with this way come two new challenges.

\textbf{Challenge 1: training environment dilemma.} Training agents in simulated environments is sample-efficient, yet ensuring their realism remains challenging.   
For instance, previous research in RL-based autonomous pentesting \cite{schwartz2019autonomous,zennaro2020modeling,Hu2020AutomatedPT,Tran2021DeepHR,abs-2202-10630, Chen2023GAILPTAI, YangCFL23, 10681147} mainly focused on training agents to find attack paths within simulated environments like NASim \cite{schwartz2019nasim} or CyberBattleSim  \cite{msft:cyberbattlesim}, while these environments are highly abstract and unrealistic as they assume that the network environment can be observed from a global perspective before training, which is unrealistic for pentesting in the real world \cite{AprilAE}.
By contrast, emulated environments (e.g., virtual machines) closely mimic real-world settings, but training agents in them is sample-inefficient and time-consuming.  To tackle this challenge, one can improve the agents' learning efficiency in emulated environments or construct simulated environments close to real-world settings.

\textbf{Challenge 2: poor generalization ability.} That is, even with suitable training environments, agents' learned policies often exhibit poor performance when transferred to unseen testing (real-world) scenarios  \cite{WangKSF20}.   The worst of it is even a slight change in the environment would require the agent's policy to be retrained.  This phenomenon is known as the \textit{generalization gap} \cite{SongJTDN20, KirkZGR23},  which significantly constrains the widespread applicability of pentesting agents.   To tackle this, one needs to improve agents' generalization ability.  
The reasons for the generalization gap are twofold: 
\begin{enumerate}
    \item RL algorithms tend to overfit the training environments \cite{CobbeKHKS19};
    \item Training environments have limited diversity, whereas real-world environments are unknown for agents and unpredictably diverse. The differences between the training and real-world environments are collectively known as the\textit{ reality gap} \cite{TobinFRSZA17}.
    
\end{enumerate}

\begin{figure*}[htbp]
	\centering
	\subfloat[Reality gap: hosts with the same vulnerability have different configurations.]
	{\includegraphics[width=0.45\textwidth]{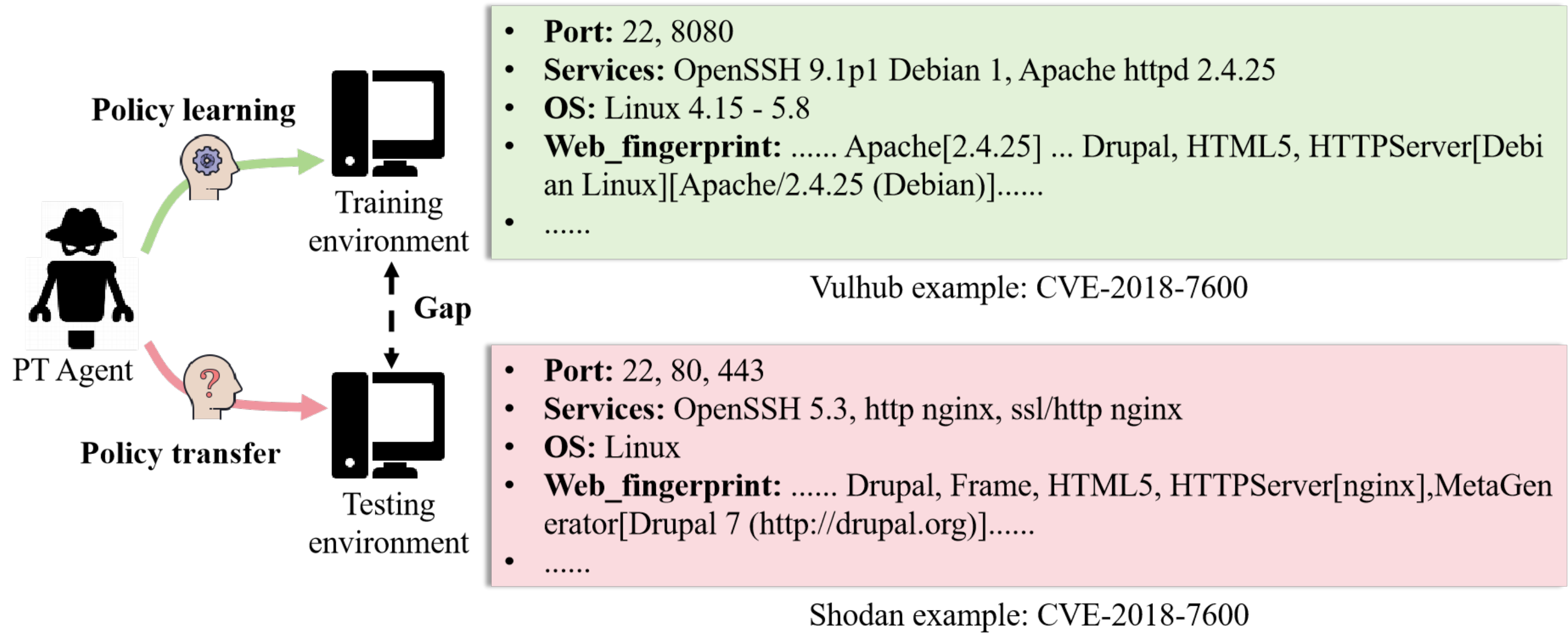}\label{fig:example_gap}}
	\quad
	\subfloat[Generalization gap between training and testing environments.]
{\includegraphics[width=0.3\textwidth]{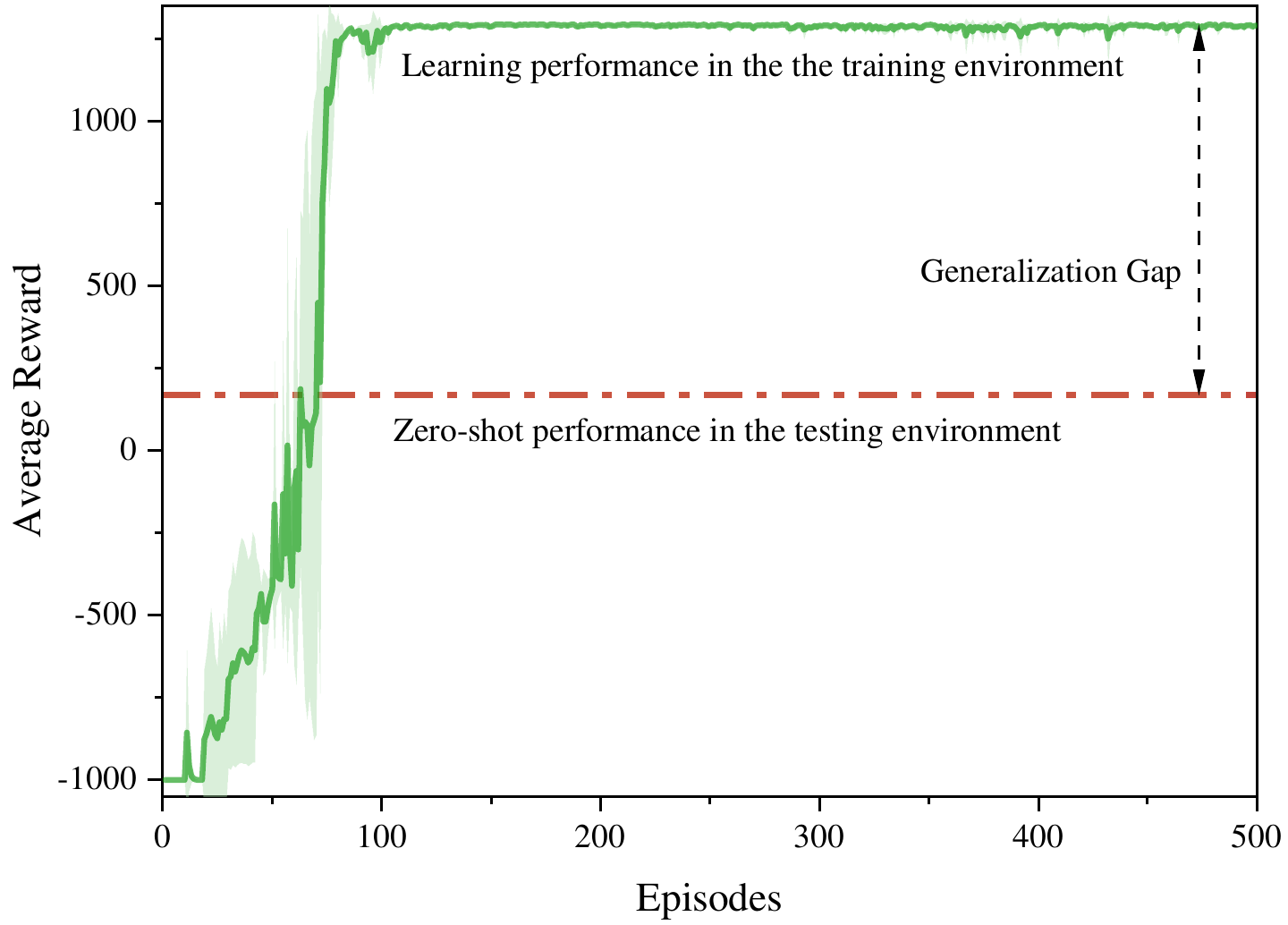}\label{fig:gengap}}
	\caption{Gap between the training and testing environments. \textbf{(a)} Shows the reality gap between training and testing environments. The training environment is a virtual machine with the CVE-2018-7600 vulnerability, set up using Vulhub\protect\footnotemark[1]. We utilized the Shodan\protect\footnotemark[2] engine to search real-world hosts that may possess the same vulnerability, which is assumed to be a testing environment. Even though both training and testing environments have the same vulnerability, differences in host configurations create a gap, thereby affecting the agent's transfer learning performance.  \textbf{(b)} Shows the learning curve of April in the training environment and its zero-shot performance in the testing environment.  The averages and 95\% confidence intervals (shaded areas) across 3 random seeds are shown.}
    \label{fig:example}
\end{figure*}

Fig.\ref{fig:example} presents an example of the reality gap and generalization gap encountered in the pentesting field.
As the example depicted in Fig.\ref{fig:example_gap}, even if vulnerabilities are the same between training and real-world testing environments, the configurations of hosts can differ. For instance, vulnerable products may be exposed on various ports, hosts could be running different operating systems, and there might also be numerous other ports open that are unrelated to the vulnerabilities, as well as potentially different web fingerprints.  These differences create a gap between these environments, leading to the agent perceiving different observations.
Then, a generalization gap will occur once the agent learns to rely on certain observational features in the training environment that change in the testing environment. 
Fig. \ref{fig:gengap}  visually demonstrates the impact of the generalization gap on the agent's policy transfer. We train the agent using April \cite{AprilAE} (the state-of-the-art framework for autonomous pentesting) in the training environment shown in Fig. \ref{fig:example_gap}, and then transfer the trained agent to the testing environment for evaluation. The zero-shot performance in the testing environment shows a significant decline compared to the learning performance in the training environment.
We further demonstrate this phenomenon in Section \ref{exp2}, especially in Fig.\ref{fig: exp2}.


We argue that an agent with good generalization ability should be able to draw inferences about other cases from one instance, akin to human capabilities.
Specifically, when the testing environments are similar to the training environment, i.e., have the same vulnerabilities, the agent should be able to bridge the generalization gap and achieve zero-shot policy transfer.  Moreover, it should also demonstrate the capacity for few-shot policy adaptation in dissimilar scenarios featuring varying vulnerabilities, thereby enhancing overall learning efficiency.

To tackle the two challenges, we propose a {\textbf{G}}eneralizable {\textbf{A}}utonomous {\textbf{P}}entesting framework, namely \textbf{GAP}. 
GAP works following a "Real-to-Sim-to-Real" pipeline, where the training environment dilemma and poor generalization ability can be tackled.
In the \textbf{Real-to-Sim phase}, GAP realizes end-to-end learning in unknown real/emulated environments without requiring significant human effort, while constructing realistic simulation analogs for real environments. 
On this basis, in the \textbf{Sim-to-Real phase}, GAP aims to enhance agents' generalization ability, enabling them to achieve two key objectives: (a) bridging the generalization gap for zero-shot policy transfer in similar scenarios, and (b) facilitating rapid policy adaptation in dissimilar scenarios to enhance learning efficiency. For this purpose, two methods are applied:

\textbf{1. Environment augmentation:}  
Similar to data augmentation in supervised or unsupervised learning, domain randomization \cite{ChenHJLW22, HorvathEIHF23} can be considered a form of environment augmentation technique that is widely used in robotics to improve the robustness and generalization of models by highly randomizing the rendering settings for the simulated training set.  
Besides, large language models (LLMs) have recently demonstrated an unprecedented ability 
to generate synthetic data for specific tasks \cite{LiZL023, abs-2403-04190}, providing us with a promising approach to domain randomization. 
Inspired by these, 
this paper proposes an LLM-powered domain randomization method to synthesize sufficient simulated training environments based on limited real-world data. 
This approach offers a feasible solution to increase both the amount and diversity of training environments. 
By employing domain randomization to randomly change part of the host configuration during training, the agent can prevent its policy from overfitting to specific observational features, thereby enhancing policy robustness. 

\textbf{2. Learning to learn: }
Meta-learning \cite{HospedalesAMS22, YeZGZ24}, also known as learning to learn,  aims to enable models to adapt to unseen tasks quickly by leveraging prior learning experiences on multiple training tasks. Meta-RL is to apply this principle to RL. 
Recent advancements highlight its potential to address RL's inherent overfitting and sample inefficiency \cite{abs-2301-08028, FinnAL17, Botvinick2019ReinforcementLF}.
While Meta-RL facilitates efficient adaptation of agents to unseen testing environments, it necessitates extensive meta-training across diverse related tasks. However, due to the absence of appropriate training environments, Meta-RL has not been successfully applied in autonomous pentesting before. 
To address this challenge, this paper initially employs domain randomization to create an adequate number of simulated training environments. Subsequently, it applies meta-RL to extract inductive biases from these synthetic environments, thereby enhancing the agents' generalization ability.

To sum up, our main contributions are threefold:
\begin{enumerate}
	\item We propose GAP, a generalizable autonomous pentesting framework that works on a Real-to-Sim-to-Real pipeline. This framework enables end-to-end policy learning in unknown real environments as well as the construction of realistic simulations, while improving agents' generalization ability.
	\item To bridge the generalization gap and realize fast policy adaptation, we are among the first to apply domain randomization in the autonomous pentesting domain and propose an LLM-powered domain randomization method for environment augmentation.  We further apply meta-RL to improve the agents' generalization ability to unseen environments by leveraging the generated simulations.
	\item We conduct experiments on various vulnerable virtual machines. The experimental results demonstrate that our framework can (a) enable policy learning in various realistic  environments, (b) achieve zero-shot policy transfer in similar environments, and (c) realize rapid policy adaptation in dissimilar environments.
\end{enumerate}

\footnotetext[1]{https://github.com/vulhub/vulhub}
\footnotetext[2]{https://www.shodan.io}

\section{Related Work}

\subsection{Autonomous Penetration Testing}

Research on autonomous pentesting originated from automated attack planning, which used planning algorithms, such as attack trees and attack graphs  \cite{schneier1999attack, Sheyner2002AutomatedGA, JmalHBIKW24},  to discover possible attack paths in target networks.   
Planning algorithms offered interpretable and formal models for evaluating network security.  However, both approaches rely on unrealistic assumptions that complete knowledge of the target network topology and hosts’ configurations are known.

Recently, RL has shown remarkable progress in various domains, including video games \cite{Vinyals2019GrandmasterLI}, autonomous vehicles \cite{Feng2023DenseRL}, unmanned aerial vehicles \cite{BoZZHLZL24}, and cybersecurity \cite{LiLYP22}.   
Compared to traditional planning methods, RL allows for training an agent to learn a policy through trial and error while interacting with the environment, eliminating the need for supervision or environmental models. 
This agent-environment interaction learning paradigm makes RL  suitable for studying autonomous pentesting.   

Previous research on RL-based autonomous pentesting mainly focused on training agents to discover attack paths in simulated training environments like NASim \cite{schwartz2019nasim} or CyberBattleSim \cite{msft:cyberbattlesim} that are highly abstract and unrealistic.
They made attempts to improve the learning efficiency of the pentesting agents through various methods, such as action space decomposition \cite{Tran2021DeepHR},  distributed algorithm \cite{IlicDVMP24}, or the introduction of demonstration data \cite{Chen2023GAILPTAI} and curiosity mechanism \cite{YANG:193309}. Besides,  \cite{YangCFL23} and \cite{10681147} have provided solutions to enhance the adaptability of agents in response to dynamic changes in such simulated environments.
Remarkable progress has been made, though, they heavily rely on an over-optimistic and unrealistic assumption that the network environment, including topologies, host configurations, and even vulnerabilities, can be observed from a global perspective, which is unrealistic even for human experts \cite{AprilAE}.

There has also been some research \cite{de, Maeda2021AutomatingPW, AprilAE} attempts to train pentesting agents in more realistic environments.  
However, these efforts often trained agents to learn policies in  fixed scenarios. Such learned policies frequently suffer from poor generalization, as they tend to overfit specific training scenarios or tasks.  Consequently, even a slight change in the environment would require agents' policies to be retrained.
This limitation hinders their adaptability to the diverse vulnerability environments encountered in the real world, posing challenges to the widespread deployment of pentesting agents. 

Different from previous research, this work follows the settings of \cite{AprilAE}, with the goal of 
 training pentesting agents that can be generalized to unseen real-world scenarios.  This generalization ability refers to drawing inferences from one instance to another, enabling agents to bridge the generalization gap for zero-shot policy transfer in similar environments with the same vulnerabilities, as well as achieve few-shot policy adaptation in dissimilar scenarios.

\subsection{Domain Randomization}
Domain randomization is  a simple but effective technique widely used in vision-based robot control tasks, like fetch robot \cite{TobinFRSZA17} and vehicles \cite{8575297}, to enhance the robustness and generalization of models \cite{HorvathEIHF23,PengAZA18,TiboniPTA23}. 
Its key insight is that with enough variability in the simulator, the real world may appear to the agent as just another variation \cite{TobinFRSZA17}.   
It achieves this by randomizing various parameters related to the simulated training environment, such as rendering settings, textures, lighting conditions, physics properties, and object characteristics. This variability helps the models adapt to a wide range of conditions during training, thereby improving their ability to perform effectively in real-world scenarios where conditions may vary unpredictably.  

Currently, domain randomization has not yet been applied in autonomous pentesting.
Different from these vision-based tasks, autonomous pentesting agents observe environments using various scanning tools, which typically provide feedback in textual format. 
Compared to visual data, text data involves semantic, syntactic, and logical structures of language rather than straightforward physical features. This abstract nature makes randomizing text data more complex because any variation must maintain semantic coherence, specialization, and intelligibility, especially in a specialized field like pentesting.

In supervised learning and unsupervised learning, data augmentation can increase both the amount  and the diversity  of a given dataset \cite{ChenTRBY23}, while domain randomization can  be considered a form of scenario augmentation technique.   In recent years, generative AI, particularly LLMs, has made significant progress in both text comprehension and generation.  LLMs provide a mechanism to create rich, contextually relevant synthetic data on an unprecedented scale \cite{abs-2403-04190, long2024llmsdrivensyntheticdatageneration}, and also offer a new solution to data augmentation \cite{abs-2402-14568,abs-2403-02990,abs-2305-14288}.  Inspired by this, this paper takes the first step to applying domain randomization in autonomous pentesting, and proposes an LLM-powered scenario augmentation method to increase both the amount  and the diversity  of training environments, thereby improving agents' generalization ability.

\subsection{Meta-Reinforcement Learning}
There are two main kinds of meta-RL methods that emerge from meta-learning: gradient-based methods and recurrence-based methods.  
Gradient-based methods try to learn initial neural network parameters that are applicable across multiple tasks and are easily fine-tuned. This enables the model to achieve optimal performance with minimal data from new tasks, using just one or a few gradient steps to update the parameters.
In contrast, recurrence-based methods leverage recurrent neural networks to enhance the model's capacity for leveraging historical experiences \cite{JuJGNL22}.   The representative algorithms for these methods are Model-Agnostic Meta-Learning (MAML) \cite{FinnAL17} and RL$^{2}$ \cite{DuanSCBSA16}, respectively, with other algorithms expanding upon similar concepts and techniques derived from these foundational approaches.

In this paper, we apply MAML to perform meta-training. Compared to recurrence-based methods, gradient-based methods have advantages in parameter efficiency, computational efficiency, and new task adaptation. They can also be combined with many policy gradient-based RL algorithms easily since the inner-loop optimization process is just a policy gradient algorithm \cite{ArndtHGK20}. 
Furthermore, under certain assumptions, gradient-based methods can enhance policy performance from the initialization on any given task, even one from outside the task distribution \cite{abs-2301-08028}.

\section{Preliminary}

\subsection{Reinforcement Learning}

RL is a machine learning method that maps the state of the environment to actions \cite{sutton2018reinforcement}. Markov Decision Processes (MDPs) serve as the mathematical foundation for RL algorithms, providing a formal framework for modeling sequential decision-making.  

A discrete-time MDP can be formalized by a tuple $(\mathcal{S},\mathcal{A},\mathcal{R},\mathcal{P} ,\gamma)$, where $\mathcal{S}$ represents the state space, $\mathcal{A}$ is the action space, $\mathcal{R}:\mathcal{S} \times \mathcal{A} \mapsto \mathbb{R} $ is the reward function, $\mathcal{P}:\mathcal{S} \times \mathcal{A}\times \mathcal{S}\mapsto [0,1]$ is the state transition probability function and $\gamma$ is the discount factor used to determine the importance of long-term rewards.   For a standard RL problem, at each time step $t$, the RL agent observes the state $s_{t}$ from the environment and selects an action $a_{t}$ based on its policy $\pi$. 
Then, the agent receives a reward signal from the environment, and the environment transitions to the next state $s_{t+1}$.   
The goal of the RL agent is to learn the optimal policy $\pi^{*}$ with parameters $\phi^{*}$ that maximizes the expected discounted reward within an episode. Thus, the RL objective function is defined as:
\begin{equation}
	J(\pi_{\phi})=\mathbb{E} _{\tau  \sim P_{\pi_{\phi} }}\left [ \sum_{t=0}^{T}\gamma ^{t}r_{t} \right ] ,
\end{equation}
where  $\gamma \in [0,1]$ is the discount factor used to determine the importance of long-term rewards, $T$ is the horizon, $\tau=\left \{ s_{t},a_{t},r_{t} \right \} _{t=0}^{T}$ is the $T$-step trajectory, and $P_{\pi_{\phi} }$ denotes the distribution of $\tau$ under policy $\pi_{\phi}$.  

\subsection{Meta-Reinforcement Learning}
Meta-learning, often framed as learning to learn, aims to learn a general-purpose learning algorithm by leveraging a set of tasks with a shared structure in the training phase, so that it can generalize well to new and similar tasks \cite{HospedalesAMS22, BingLHK23}.
Meta-RL, in short, is to apply this concept to RL.

Generally, considering the parameterized RL policies with parameters $\phi \in \Phi$, we can define 
an RL algorithm as the function $f$ that maps the training data to a policy: $f(D )= ( \left ( \mathcal{S} \times \mathcal{A} \times \mathbb{R} \right )^{T}  )^{H} \to \Phi$, where $H$ denotes the total training episodes and $D=\left \{ \tau ^{h} \right \} _{h=0}^{H}$ is all of the trajectories collected in the MDP.  
The main idea of meta-RL is instead to learn the RL algorithm $f$ that outputs the parameters of RL policy \cite{abs-2301-08028}:  $\phi=f_{\theta }(\mathcal{D} )$, where $\theta$ is the meta-parameters need to learn to maximize a meta-RL objective, and $\mathcal{D}$ represents the meta-trajectories that may contain multiple policy trajectories.

Meta-RL usually involves two loops of learning: an inner loop, where the agent adapts its policy to a specific task, and an outer loop,  also referring to the meta-training, where the agent updates its meta-parameters $\theta$ based on the performance across multiple tasks to allow the agent to quickly adapt or learn efficiently in new tasks.  
Meta-training requires access to a set of training tasks/environments, each of which is formalized as an MDP and comes from a distribution denoted $p(\mathcal{M})$.  In this paper, the distribution is defined over different variations (e.g., different host configurations) of the pentesting tasks, which differ only in state transition probability while maintaining consistency in $\mathcal{S}$, $\mathcal{A}$, and $\mathcal{R}$.  Thus, the meta-training can be described as a process that proceeds by sampling a task from  $p(\mathcal{M})$, executing the inner loop on it, and optimizing the inner loop to improve policy adaptation.  
Formally, following \cite{abs-2301-08028}, we define the meta-RL objective  as:
\begin{equation}
	\mathcal{J} (\theta )=\mathbb{E} _{\mathcal{M} _{i}\sim p(\mathcal{M} )}\left [ \mathbb{E} _{\mathcal{D} } \left [ \sum_{\tau \in \mathcal{D} }G(\tau) \bigg|  f_{\theta},\mathcal{M}_{i}   \right ]  \right ] , \label{meta-rl objective}
\end{equation}
where $G(\tau)=\sum_{t=0}^{T}\gamma ^{t}r_{t} $ is the discounted reward in task $\mathcal{M}_{i}$, $\tau$ is the trajectory sampled from task $\mathcal{M}_{i}$ under the policy with parameters $\phi=f_{\theta }(\mathcal{D} )$.

\subsection{Generalization in Reinforcement Learning}
To evaluate the generalization ability of pentesting agents, we train the agent on a set of environments $\mathcal{M} _{train}=\left \{ \mathcal{M} _{1},...,\mathcal{M} _{n} \right \} $ with $\mathcal{M} _{i}\sim p(\mathcal{M})$, and then evaluate its generalization performance on the testing environments $\mathcal{M} _{test}$ drawn from $p(\mathcal{M})$.   

We evaluate the agents' generalization ability in two ways. 
Firstly, following \cite{KirkZGR23, LyleRDKG22}, we study the  zero-shot policy transfer performance in testing environments using the generalization gap, which is defined as 
\begin{equation}
	\mathrm{GenGap} (\pi)=\mathbb{E} _{\mathcal{M}_{train} ,\tau  \sim P_{\pi }}\left [ G(\tau) \right ] -\mathbb{E} _{\mathcal{M}_{test} ,\tau  \sim P_{\pi }}\left [ G(\tau) \right ],\label{equ: gengap}
\end{equation}
where $G(\tau)$ is the expected cumulative reward over a trajectory,  $\mathbb{E} _{\mathcal{M}_{test} ,\tau  \sim P_{\pi }}\left [ G(\tau) \right ]$ is the zero-shot performance.  This metric measures the agent's ability to overcome overfitting and achieve zero-shot generalization. A smaller generalization gap indicates that the deployment performance won’t deviate as much from the training performance, potentially indicating a more robust policy. 
Besides,  following \cite{KirkZGR23, TaigaAFCB23}, we also study  the  few-shot policy adaptation performance by allowing the agent to interact with the testing environment online. This metric can evaluate how well the agent can quickly adapt its learned policy to new tasks or environments with stronger forms of variation.

\section{Proposed Method}

\subsection{Overview}
\begin{figure*}[htbp]
	\centering
	\includegraphics[width=0.65\textwidth]{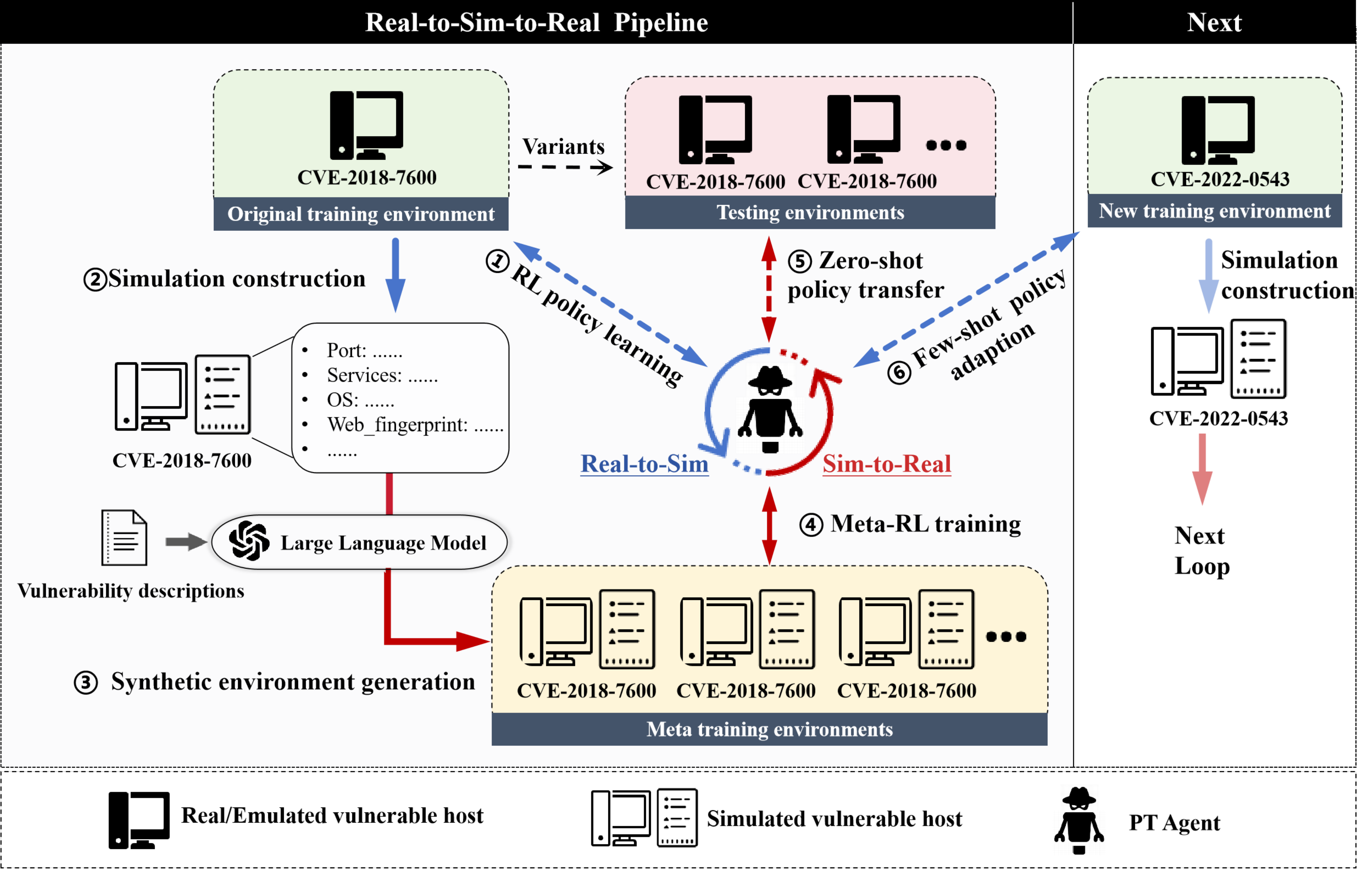}
	\caption{Overview of GAP. }
	\label{fig: gap}
\end{figure*}

GAP aims to realize efficient policy training in realistic environments and  train generalizable agents capable of drawing inferences about other cases from one instance.
"Learning from one instance" is the foundational skill, while "generalizing to another" is the core objective. 
To achieve this goal, it employs a Real-to-Sim-to-Real pipeline that integrates end-to-end policy learning, the construction of realistic simulation environments, and the enhancement of the agent's generalization ability. The full workflow of GAP is illustrated in Fig. \ref{fig: gap}.

\subsubsection{Real-to-Sim}

The increasing emergence of vulnerabilities, coupled with the complex and dynamic nature of real-world network environments, necessitates the ability of pentesting agents to learn and adapt policies in an end-to-end manner within unknown settings. 
GAP achieves this through the Real-to-Sim stage (see Section \ref{sec: real-to-sim} for details), as shown by the blue lines in Fig.\ref{fig: gap}, that is: \ding{172} end-to-end policy learning in the original training environment, and \ding{173} realistic simulation construction based on information gathered during policy learning. This constructed simulation serves as a digital mapping of the real environment, facilitating efficient policy training and environment augmentation in the subsequent stage. 

\subsubsection{Sim-to-Real}
Sim-to-Real is a comprehensive concept that has been applied to robotic and classic machine vision tasks \cite{ZhaoQW20}. The goal of sim-to-real transfer is to ensure that the behaviors, actions, or decisions learned in simulation can effectively and reliably be applied in real-world scenarios.  
Similarly, in the Sim-to-Real stage, GAP utilizes the simulated environment constructed in the last stage to improve the agent's generalization ability, allowing the agent to quickly adapt to diverse real-world settings.  
As shown by the red lines in Fig.\ref{fig: gap}, GAP achieves this through two-phase steps: \ding{174} synthetic environment generation via LLM-powered domain randomization (see Section \ref{sec: dr}), as well as \ding{175} meta-RL training based on the synthetic environments (see Section \ref{sec: meta}).  In this way, the agent is expected to achieve zero-shot policy transfer in similar environments (step \ding{176} in Fig.\ref{fig: gap}) and few-shot policy adaptation in dissimilar environments (step \ding{177} in Fig.\ref{fig: gap}).

Note that in our setup, we utilize virtual machines as surrogate real-world environments to train and test pentesting agents, since they provide controlled, isolated, and high-fidelity environments for agents to practice and refine offensive security techniques.  The testing environments are variants of the original training environment, featuring the same vulnerability but different host configurations. 

In order to make a clear distinction between different environments in this paper, we use the symbol $\mathcal{M}^{k}_{h}$ to denote a host $h$ with vulnerability $k$.  Thus, the original training environment is denoted as $\mathcal{M}^{k}_{0}$,  the simulation of $\mathcal{M}^{k}_{0}$ is denoted as $\widetilde{\mathcal{M} }^{k} _{0}$, and the set of generated synthetic environments is denoted as $\widetilde{\mathcal{M} }^{k} =[\widetilde{\mathcal{M} }^{k}  _{1},\widetilde{\mathcal{M} }  ^{k}_{2},...]$.  To simplify the symbolic representation, we omit the vulnerability identifier $k$ in subsequent sections.

\subsection{Policy Learning and Simulation Construction}\label{sec: real-to-sim}

\begin{figure}[htbp]
	\centering
	\includegraphics[width=0.4\textwidth]{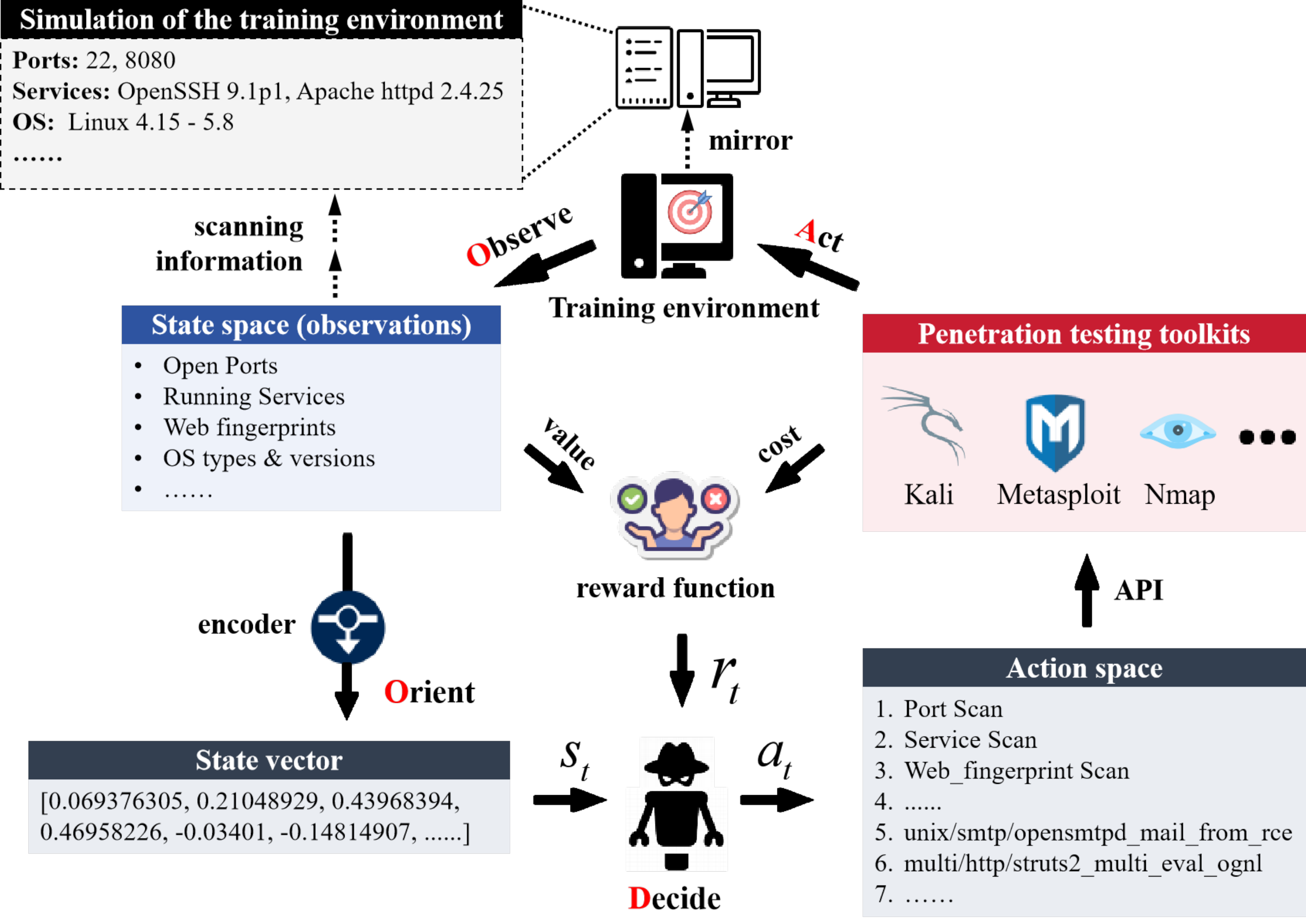}
	\caption{The process of policy learning and simulated environment construction.}
	\label{fig:agent}
\end{figure}

Being able to learn policy in unknown environments is a fundamental skill for pentesting agents, where unknown environments refer to scenarios where the agent lacks prior knowledge of the target hosts' configuration and vulnerabilities.  In this part, we introduce how we model the pentesting process as an MDP, thereby employing RL to train the agent to learn how to explore and exploit vulnerabilities using observed environmental states in an end-to-end manner.  Additionally, upon detecting host configuration details, the agent constructs simulated environments to enhance its policy generalization capabilities in subsequent stages. 
The process of policy learning and simulated environment construction is depicted in Fig.\ref{fig:agent}.

\subsubsection{Model penetration testing as an MDP}

To train the pentesting agent using RL algorithms, we model the pentesting process as an MDP following \cite{AprilAE}. This is formalized by defining the state space $\mathcal{S}$, action space $\mathcal{A}$, and reward function $\mathcal{R}$. For model-free RL algorithms, the transition probability function $\mathcal{P}$ remains unknown. 

$\mathcal{S}$ is all potential environmental states observable by the agent. 
These observations typically involve host configurations detectable using scanning tools, including scanning information like ports, services, operating systems, website fingerprints, etc. 
Human experts can use this information to pinpoint potential vulnerabilities within the target host, and thus,  these details can form the foundation for the agent to make decisions.

$\mathcal{A}$ represents the set of all actions available to the agent. 
For pentesting tasks, these actions usually include various forms of information gathering, system probing, and vulnerability exploitation. 
In contrast to previous research that used abstract actions, our approach emphasizes the use of concrete and executable actions for end-to-end autonomous pentesting. These actions are aligned with specific commands or payloads derived from real pentesting tools or systems, such as Kali\footnotemark[3], Metasploit\footnotemark[4], Nmap\footnotemark[5], and so on. 
Note that a larger action space  necessitates more extensive exploration for agents to discover effective actions, thereby  potentially reducing the tractability of the agent’s training process. This phenomenon can be seen in experiments in Section \ref{exp1}.

$\mathcal{R}$ defines the agent's learning goal. 
In our study, the agent aims to gather critical information from the target host to accurately pinpoint the potential vulnerability and exploit it to compromise the host. Throughout this process, the agent should minimize the overall cost of actions to maintain covert. For this purpose,  we define the reward function as
\begin{equation}
	\mathcal{R}=\mathrm{value}  (h)-{\textstyle \sum_{a\in A }^{}} \mathrm{cost} (a)  ,
	\label{reward_function}
\end{equation}
where $A\subseteq \mathcal{A} $ is the set of actions used in the pentesting process, $\mathrm{value}  (h)$ returns a positive reward value if the target host $h$ is compromised by exploiting the correct vulnerability or the agent successfully gains some kind of information about the host, $\mathrm{cost} (a)$ refers to the cost of action $a$.

\footnotetext[3]{https://www.kali.org/}
\footnotetext[4]{https://www.metasploit.com/}
\footnotetext[5]{https://nmap.org/}

At its core, the process of interaction between the agent and its environment can be likened, to some extent, to an \textit{Observe}, \textit{Orient}, \textit{Decide}, and \textit{Act} (OODA) loop.  This process can be described as follows.

Firstly, the agent utilizes scanning tools to \textit{observe} environmental states and collect raw observations, typically in textual format. This data often contains significant redundancy and cannot be directly fed into the agent's policy model.  
Therefore, the agent needs to \textit{orient} the raw observations by analyzing and synthesizing the observed information to construct a state vector that the neural network can comprehend. To achieve this,  we utilize the pre-trained Sentence-BERT \cite{wang-2021-TSDAE} model as the encoder for embedding the raw observations.  These embedded representations of state vectors serve several purposes: (a) capturing essential features of the raw observations in a latent space, (b) facilitating end-to-end learning, and (c) preventing the dimensional explosion of the state space. 
Thus,  the agent takes the state vector as input, makes a \textit{decision}, and then outputs an action from the action space.
Based on the output action, the agent \textit{acts} on the target host by invoking and executing the corresponding tools in pentesting toolkits via the application programming interface (API).  Finally, the interaction process transitions into the next OODA loop. 

\subsubsection{Simulated environment construction}
During interaction with the original training environment ${\mathcal{M} }_{0}$, the agent uses gathered raw observations—specifically, host configuration data—to build a simulated environment $\widetilde{\mathcal{M} }_{0}$ in JSON format. 
This simulated environment faithfully mirrors the original training environment, as it is constructed based on actual feedback data. 
Thus, theoretically, the agent interacting with this simulated environment equates to interacting with a real environment. More importantly, this simulated environment can be used not only for efficient training and validating the agent's policies but also as a real-world environment example for subsequent environment augmentation. 

\subsubsection{Policy training}
We train the agent's policy using the PPO algorithm \cite{PPO} on the original training environment $\mathcal{M}_{0}$. It is noteworthy that GAP is not limited to PPO. A wide range of policy gradient-based RL algorithms can be used out of the box.

PPO is an on-policy actor-critic method with two primary variants of PPO: PPO-Penalty and PPO-Clip, where we focus on the most widely used variant PPO-Clip. 
By using PPO, the agent updates policy $\pi_{\phi}$ by taking multiple steps of (usually minibatch) SGD to optimize the following  objective:
\begin{equation}
	\mathcal{L}^{PPO}(\phi )=\mathbb{E} _{ \tau_{t} \sim \pi }\left [ \min \left ( r_{t}(\phi) \hat{A}_{t},\mathrm{clip}\left (r_{t}(\phi),1-\epsilon ,1+ \epsilon \right )\hat{A}_{t}   \right )  \right ] ,
	\label{ppo}
\end{equation}
where $r_{t}(\phi)=\frac{\pi_{\phi}(a_{t}|s_{t})}{\pi_{\phi_{\mathrm{old}} }(a_{t}|s_{t})} $ denote the probability ratio,  $\hat{A}_{t}$ is an estimator of the advantage function which describes how much better the action is than others on average,  $\mathrm{clip}\left (r_{t}(\phi),1-\epsilon ,1+ \epsilon \right )$ modifies the surrogate objective by clipping the probability ratio, and $\epsilon$ is a small hyperparameter which roughly measures how far away the new policy is allowed to go from the old.

\subsection{LLM-powered Domain Randomization}\label{sec: dr}
In the real world, hosts with the same vulnerability are affected by the same vulnerable product (VP), but their configurations can also differ significantly (see Fig.\ref{fig:example}). They may run different versions of VP or services, have varied open ports, operate on different operating systems, or exhibit different website fingerprints. 
These differences create the reality gap, meaning that an agent trained in a single environment would rely on specific observations and thereby may struggle to directly transfer its learned policies to real-world deployment. Therefore, considering and simulating these real-world differences during training is crucial to enhancing the generalization ability of pentesting agents.

In this paper, we employ domain randomization as an environment augmentation technique, which can be feasible to address the challenge posed by limited diversity in training environments.     Unlike vision-based tasks, where domain randomization is commonly applied, pentesting agents gather information through scanning tools that provide textual feedback. Textual data encompasses semantic, syntactic, and logical structures of language rather than straightforward physical attributes. This abstract nature complicates the randomization of text data, as any variation must preserve semantic coherence, specificity, and comprehensibility. 
The recent advancements in LLMs demonstrate the capability to generate synthetic data that mimics the characteristics and patterns of real-world data, thereby being a suitable and promising approach to performing domain randomization for autonomous pentesting.

\begin{figure}[htbp]
	\centering
	\includegraphics[width=0.4\textwidth]{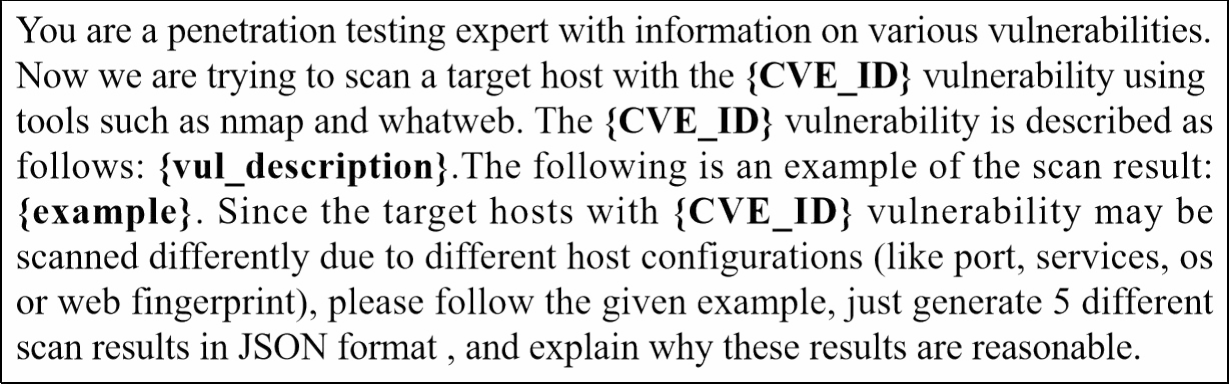}
	\caption{Prompt pattern for synthetic environment generation.}
	\label{fig:prompt}
\end{figure}

Vulnerability descriptions in the NVD repository\footnotemark[6] detail key characteristics such as VPs, affected versions, product vendors, and impact. The constructed simulation of the original training environment serves as a realistic example. Therefore, we use these descriptions as background knowledge and the constructed simulation as task examples to design prompts for LLMs.   The prompt pattern is shown in Fig.\ref{fig:prompt}. 
\footnotetext[6]{https://nvd.nist.gov/}

\begin{figure}[htbp]
	\centering
	\includegraphics[width=0.45\textwidth]{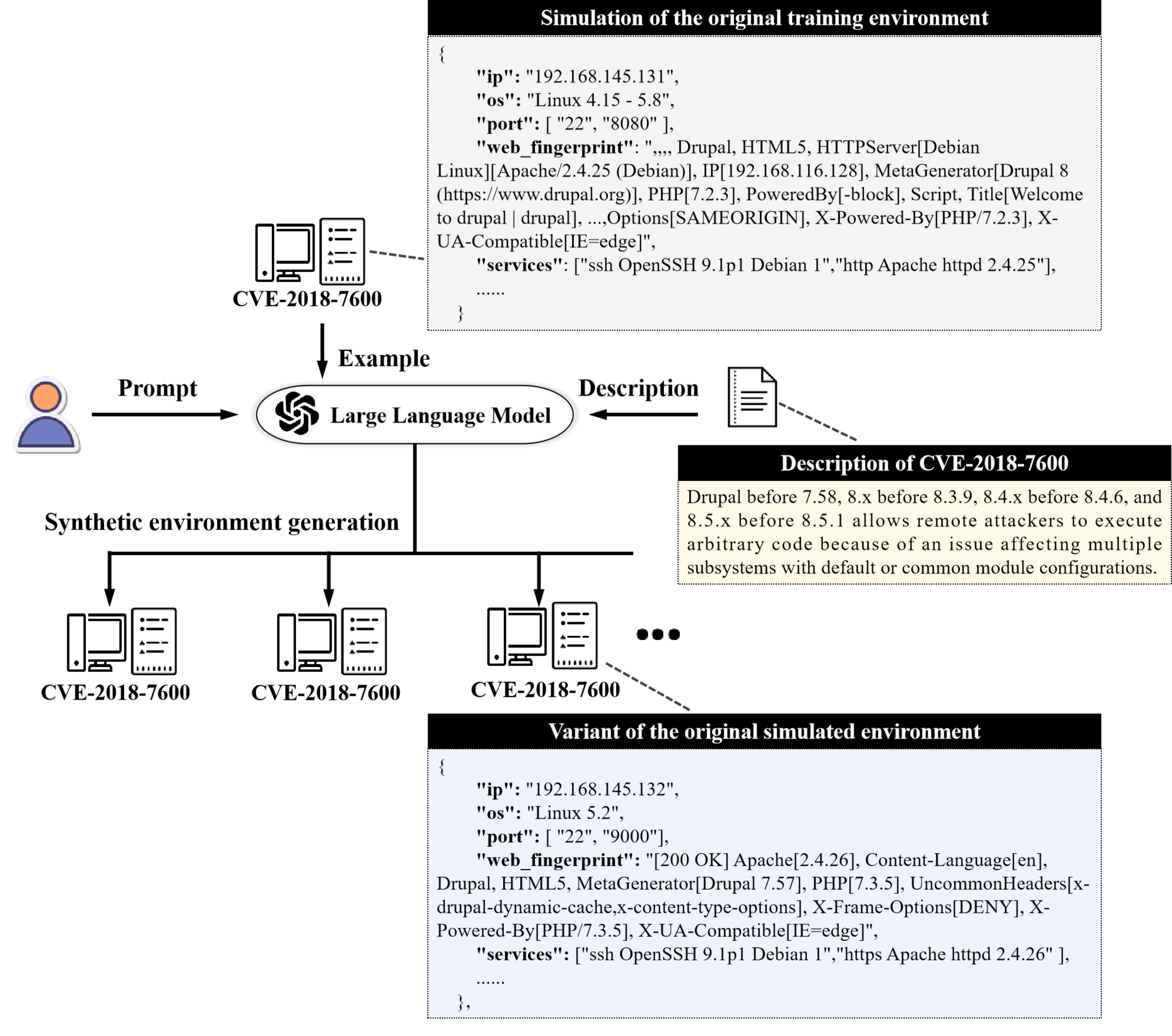}
	\caption{The workflow and example of synthetic environment generation using LLM. In this example, we construct the simulation by using the CVE-2018-7600 vulnerable host from Vulhub as the original training environment. Then, we employ GLM-4 \cite{glm2024chatglm} for domain randomization.  }
	\label{fig:llm}
\end{figure}

Fig.\ref{fig:llm} provides the workflow and an example of synthetic environment generation using LLM. In this example, the LLM generates a variant of the original simulated environment based on the official vulnerability description and original simulation $\widetilde{\mathcal{M} }_{0}$. In this variant, the web application (Drupal) is exposed on a non-default port (9000), and there are random changes in the operating system version, Apache HTTP Server version, and Drupal version compared to the original simulation.  These changes achieve domain randomization, making the synthetic environment better replicate the diversity found in the real world, thereby preventing agents from relying on fixed host configuration details.

\subsection{Meta-RL Training}\label{sec: meta}

Given the set of synthetic environments $\widetilde{\mathcal{M} } =[\widetilde{\mathcal{M} }  _{1},\widetilde{\mathcal{M} }_{2},...]$ generated by the LLM on the basis of $\widetilde{\mathcal{M} }_{0}$, in this part, we leverage $\widetilde{\mathcal{M} }$ as meta-training environments and employ the MAML algorithm \cite{FinnAL17} for meta-RL training. 
It's important to note that while MAML is utilized here, a wide range of other meta-RL algorithms are also applicable.

In MAML, the inner loop is a policy gradient algorithm with initial parameters as meta-parameters $\phi = \theta$.  
The key insight of MAML is that its inner loop is a differentiable function of the initial parameters \cite{abs-2301-08028}. This means that the initialization of the model can be optimized using gradient descent to find a set of initial parameters that serve as a good starting point for learning new tasks drawn from the task distribution.  During the new task adaptation phase, MAML employs an on-policy policy gradient algorithm to update the policy parameters. This requires MAML to sample new trajectories $\mathcal{D} _{i}$ using the initial policy,  and then use these trajectories to update a set of parameters by applying a policy gradient step  for a task/environment $\widetilde{\mathcal{M} }  _{i}\in \widetilde{\mathcal{M} }  $:
\begin{equation}
	\phi '_{i}=\phi+\alpha  \bigtriangledown _{\phi}\hat{J}_{\widetilde{\mathcal{M} }  _{i}} (\pi_{\phi},\mathcal{D} _{i}), \label{eq:adaptation}
\end{equation}
where $\hat{J}_{\widetilde{\mathcal{M} }  _{i}} (\pi_{\phi},\mathcal{D} _{i})$ is an estimate of the expected discounted reward of policy $\pi_{\phi}$ for task $\widetilde{\mathcal{M} }  _{i}$, and $\alpha$ is the learning rate.  After adapting to each task, MAML collects new trajectories $\mathcal{D}' _{i}$  again using adapted policy $\pi_{\phi'_{i}}$ for updating the initial parameters $\phi$ in the outer loop:
\begin{equation}
	\phi\gets \phi+\beta  \bigtriangledown _{\phi}\sum_{\widetilde{\mathcal{M} }  _{i}\in \widetilde{\mathcal{M} } }\hat{J}_{\widetilde{\mathcal{M} }  _{i}} (\pi_{\phi'_{i}},\mathcal{D}' _{i}),\label{eq:meta-update}
\end{equation}
where $\pi_{\phi'_{i}}$ is the policy for task $\widetilde{\mathcal{M} }  _{i}$ adapted by the inner loop,  $\hat{J}_{\widetilde{\mathcal{M} }  _{i}} (\pi_{\phi'_{i}},\mathcal{D}' _{i})$ is the gradient of the expected discounted reward of the adapted policy calculated with respect to the initial parameters.
The gradient collected from the inner loop is referred to as the meta-gradient, which reflects how the agent’s performance on the new task affects the updates to the agent’s policy parameters.  Descending the meta-gradient is essentially performing gradient descent on the meta-RL objective given by Equation \ref{meta-rl objective}.  
The complete meta-RL training algorithm is presented in Algorithm \ref{alg1}.

\begin{algorithm} 
	\caption{Meta-RL Training with MAML} 
	\label{alg1} 
	\begin{algorithmic}[1]
		\REQUIRE $\widetilde{\mathcal{M} }$: meta-training environments
		\REQUIRE $\alpha, \beta$: learning rates, and $\phi$: initial policy parameters
		\FOR {each task environment $\widetilde{\mathcal{M} }  _{i}\in \widetilde{\mathcal{M} } $}
		\STATE Sample trajectories $\mathcal{D} _{i}$ using policy $\pi_{\phi}$ in $\widetilde{\mathcal{M} }  _{i}$
		\STATE Compute adapted parameters with gradient descent following Equation \ref{eq:adaptation}
		\STATE Sample new trajectories $\mathcal{D}' _{i}$ using policy $\pi_{\phi'_{i}}$ in $\widetilde{\mathcal{M} }  _{i}$
		
		\ENDFOR
		\STATE Update policy parameters  $\phi$ following Equation \ref{eq:meta-update} using each $\mathcal{D}' _{i}$
		
	\end{algorithmic} 
\end{algorithm}



\section{Experimental Settings}


\subsection{Penetration testing environments}
To avoid potentially unpredictable impacts on real-world hosts and better evaluate the performance of GAP in real scenarios,  we conduct experiments in high-fidelity vulnerability environments set up using Vulhub.  Vulhub is an open-source collection of vulnerable Docker environments that offers a flexible way to easily create various pentesting environments. 
In this way, we can balance experimental control with the validation of real-world performance.

Table \ref{tab: scenarios} lists all experimental vulnerability environments sourced from Vulhub,  each serving as an original training environment $\mathcal{M}^{k}_{0}$  in GAP, where $k$ represents the CVE identifier (CVE ID). 
We selected these vulnerability environments for training and testing the performance of the pentesting agent because they target various representative vulnerable products. And more importantly, we have reliable and stable exploits of these vulnerabilities available for the agent to use. Theoretically, GAP can be extended to a broader range of vulnerability environments.

\subsection{Agent settings}
In each environment, the agent is trained for 500 episodes with a maximum of 100 iteration steps per episode.
During the pentesting process, the agent starts in an initial state without prior knowledge of the target host. Its task is to gather information and then exploit the corresponding vulnerability to compromise the host within limited steps.  Its available actions include various information scanning tools (e.g., Nmap, whatweb, dirb, etc.) and exploits in MSF.  In this paper, we change the size of the action space $\left | \mathcal{A}  \right | $ by randomly sampling different numbers of vulnerability exploits from MSF.  

Following \cite{AprilAE}, our reward function is formulated based on subjective estimates provided by human experts. Specifically, a reward of +1000 is assigned when the agent successfully exploits the correct vulnerability and compromises the target host. Additionally, a reward of +100 is allocated for proficiently gathering information about the target, such as the OS type, version, running services, and website fingerprint. Conversely, incorrect or illogical actions result in a penalty of -10. 

\subsection{Implementation details}
All vulnerability environments are deployed on virtual machines. We train the pentesting agent on a 64-bit laptop powered by an Intel(R) Core(TM) i9-12900H CPU @ 2.50GHz, with 32GB of memory, and an NVIDIA GeForce RTX 3060 Laptop GPU. The laptop runs Kali Linux and is equipped with Metasploit Framework (MSF) version v6.4.5-dev. PyTorch is utilized as the backend for implementing GAP. 

In GAP, we employ the classical PPO algorithm for policy learning and the MAML algorithm for meta-RL training. 
The synthetic environments used for meta-RL training are generated using GLM-4 \cite{glm2024chatglm} due to its satisfactory performance, and we assess their validity based on expert evaluators who are familiar with real-world pentesting scenarios. 
In fact, as a simple and general framework, GAP is not restricted to specific RL or meta-RL algorithms, nor is it limited to particular LLMs. 

Our code, training and testing environments, and hyperparameters, are publicly available for further replicability and future research: \url{https://github.com/Joe-zsc/GAP}.

\begin{table*}[htb]
\footnotesize
\caption{ Vulnerability environments.}
	\label{tab: scenarios}
\begin{tabular}{>{\raggedright\arraybackslash}p{0.15\linewidth}>{\raggedright\arraybackslash}p{0.5\linewidth}>{\raggedright\arraybackslash}p{0.2\linewidth}}
\hline
\textbf{CVE ID} & \textbf{Vulnerability Name}                                   & \textbf{Vulnerable Product} \\ \hline
CVE-2018-7600   & Drupal Core Remote Code Execution Vulnerability               & Drupal Core                 \\
CVE-2019-0230   & Apache Struts OGNL Remote Code Execution Vulnerability        & Apache Struts               \\
CVE-2019-9193  & PostgreSQL Authenticated Remote Code Execution Vulnerability    & PostgreSQL           \\
CVE-2020-10199  & Sonatype Nexus Repository Remote Code Execution Vulnerability & Nexus                       \\
CVE-2020-7247   & OpenSMTPD Remote Code Execution Vulnerability                 & OpenSMTPD                   \\
CVE-2020-7961  & Liferay Portal Deserialization of Untrusted Data Vulnerability  & Liferay Portal       \\
CVE-2020-9496   & Apache OFBiz Remote Code Execution Vulnerability              & Apache OFBiz                \\
CVE-2020-16846  & SaltStack Salt Shell Injection Vulnerability                  & SaltStack Salt              \\
CVE-2021-22205  & GitLab Unauthenticated Remote Code Execution Vulnerability    & GitLab                      \\
CVE-2021-25646  & Apache Druid Remote Code Execution Vulnerability              & Apache Druid                \\
CVE-2021-3129   & Laravel Ignition File Upload Vulnerability                    & Ignition                    \\
CVE-2021-41773  & Apache HTTP Server Path Traversal Vulnerability               & Apache HTTP Server          \\
CVE-2022-0543   & Debian-specific Redis Server Lua Sandbox Escape Vulnerability & Redis                       \\
CVE-2022-22947 & VMware Spring Cloud Gateway Code Injection Vulnerability        & Spring Cloud Gateway \\
CVE-2022-46169  & Cacti Command Injection Vulnerability                         & Cacti                       \\
CVE-2023-21839  & Oracle WebLogic Server Unspecified Vulnerability              & Oracle WebLogic             \\
CVE-2023-32315  & Ignite Realtime Openfire Path Traversal Vulnerability         & Openfire                    \\
CVE-2023-38646  & Metabase Pre-Auth Remote Code Execution Vulnerability         & Metabase                    \\
CVE-2023-42793  & JetBrains TeamCity Authentication Bypass Vulnerability        & JetBrains                   \\
CVE-2023-46604 & Apache ActiveMQ Deserialization of Untrusted Data Vulnerability & Apache ActiveMQ      \\ \hline
\end{tabular}%
\end{table*}

\subsection{Research Questions}
GAP aims to improve the agent's generalization ability, thereby allowing the agent to draw inferences about other cases from one instance.   
We aim to answer the following research questions (RQs) from  experiments and analysis:
\begin{itemize}
	\item \textbf{RQ1 (Learn from one): } Can GAP perform policy learning in various real environments? 
	\item \textbf{RQ2 (Generalize to another):}  Can GAP bridge the generalization gap and achieve zero-shot policy transfer in similar environments?
	\item \textbf{RQ3 (Generalize to another):}  Can GAP achieve rapid policy adaptation in dissimilar environments?
\end{itemize}

\subsection{Evaluation Metrics}

Table \ref{tab:metrics} presents the training and testing environments, along with the metrics used to investigate all research questions. The details are as follows.

\subsubsection{Evaluation of RQ1}
To investigate RQ1,  in each run, we evaluate the agent's learning curve and training time for all original training environments as listed in Table \ref{tab: scenarios}.  
We compare the performance of the agent with different sizes of action spaces ($
\left | \mathcal{A}  \right |  \in {\left \{ 100,500,1000 \right \} }$). 
We implement three independent runs using different seeds for each vulnerability environment, with average results shown in Section \ref{exp1}.  

\subsubsection{Evaluation of RQ2}

To investigate RQ2,  following \cite{KirkZGR23, LyleRDKG22}, we assess the agent's zero-shot generalization performance in both training and testing environments, which measures its ability to bridge the generalization gap (GenGap, calculated by Equation \ref{equ: gengap} ).  
Besides, the average success rate is also used as a metric, which reflects the proportion of agent successfully compromised hosts to the total number of hosts in testing environments. 

To this end,   we create three variants for each $\mathcal{M}^{k}_{0}$ in Table \ref{tab: scenarios} by changing its host configurations to serve as testing environments  denoted by $
\mathcal{M}^{k}_{test}=\left \{ \mathcal{M}^{k}_{1},\mathcal{M}^{k}_{2},\mathcal{M}^{k}_{3} \right \}$. 
These variants in $\mathcal{M}^{k}_{test}$ are similar to $\mathcal{M}^{k}_{0}$ as they share the same vulnerability $k$, but they differ in host configurations such as ports, services, operating system versions, website fingerprints, and others.  These differences create reality gaps that mimic the diversity and variability found in real-world host environments.  

In each implementation run, we train the agent for each $\mathcal{M}^{k}_{0}$ in Table \ref{tab: scenarios} , and later evaluate its zero-shot performance in $\mathcal{M}^{k}_{0}$ and testing environments $\mathcal{M}^{k}_{test}$, as well as their pentesting success rate in $\mathcal{M}^{k}_{test}$.
We implement three independent runs using different seeds for each vulnerability environment, with average results shown in Section \ref{exp2}.

\subsubsection{Evaluation of RQ3}
A well-generalized agent should also be able to effectively improve its learned policy to adapt to a new environment.
To investigate RQ3, following \cite{KirkZGR23, TaigaAFCB23}, we evaluate the agent's few-shot policy adaptation performance in dissimilar environments by allowing the agent to interact with a new environment online.  

In particular, in each run,  for each $\mathcal{M}^{k}_{0}$ in Table \ref{tab: scenarios}, we randomly select another environment $\mathcal{M}^{k'}_{0}$ as the testing environment, where $k' \ne k$. 
The agent is trained in $\mathcal{M}^{k}_{0}$ and transfers its learned policy in  $\mathcal{M}^{k'}_{0}$.  
We evaluate its learning curve and training time in the testing environment, which measures how quickly the agent can adapt its performance to a new task. 

In addition, following \cite{ZhuLJZ23}, we also evaluate the agent's jumpstart performance, which refers to the agent's initial performance when it starts interacting with the testing environment. It assesses how quickly the agent can abstract broader knowledge from past experiences and apply it to novel situations, thereby adapting its policy and achieving meaningful rewards in the early stages of training.  
We implement three independent runs using different seeds for each vulnerability environment, with average results shown in Section \ref{exp3}.


\begin{table}[htbp]
\caption{ Training and testing environments and metrics for investigating research questions. }
\label{tab:metrics}
\begin{threeparttable} 
\resizebox{\columnwidth}{!}{%
\begin{tabular}{llll}
\hline
RQ  & Training & Testing & Metrics \\ \hline
RQ1 &     $\mathcal{M}^{k}_{0}$                      &               --       &   Learning curve, Training time      \\
RQ2 &     $\mathcal{M}^{k}_{0}$                  &      $
\mathcal{M}^{k}_{test}=\left \{ \mathcal{M}^{k}_{1},\mathcal{M}^{k}_{2},\mathcal{M}^{k}_{3} \right \}$                &   GenGap, Success Rate      \\
RQ3 &      $\mathcal{M}^{k}_{0}$                 &       $\mathcal{M}^{k'}_{0}, k'\ne k$                &  Learning curve, Training time, Jumpstart       \\ \hline
\end{tabular}%
}
    \end{threeparttable}
    
\end{table}

\section{Experimental Results}
\subsection{Policy Learning in Real Environments (RQ1)}\label{exp1}

\begin{figure}[htbp]
	\centering
	\includegraphics[width=0.35\textwidth]{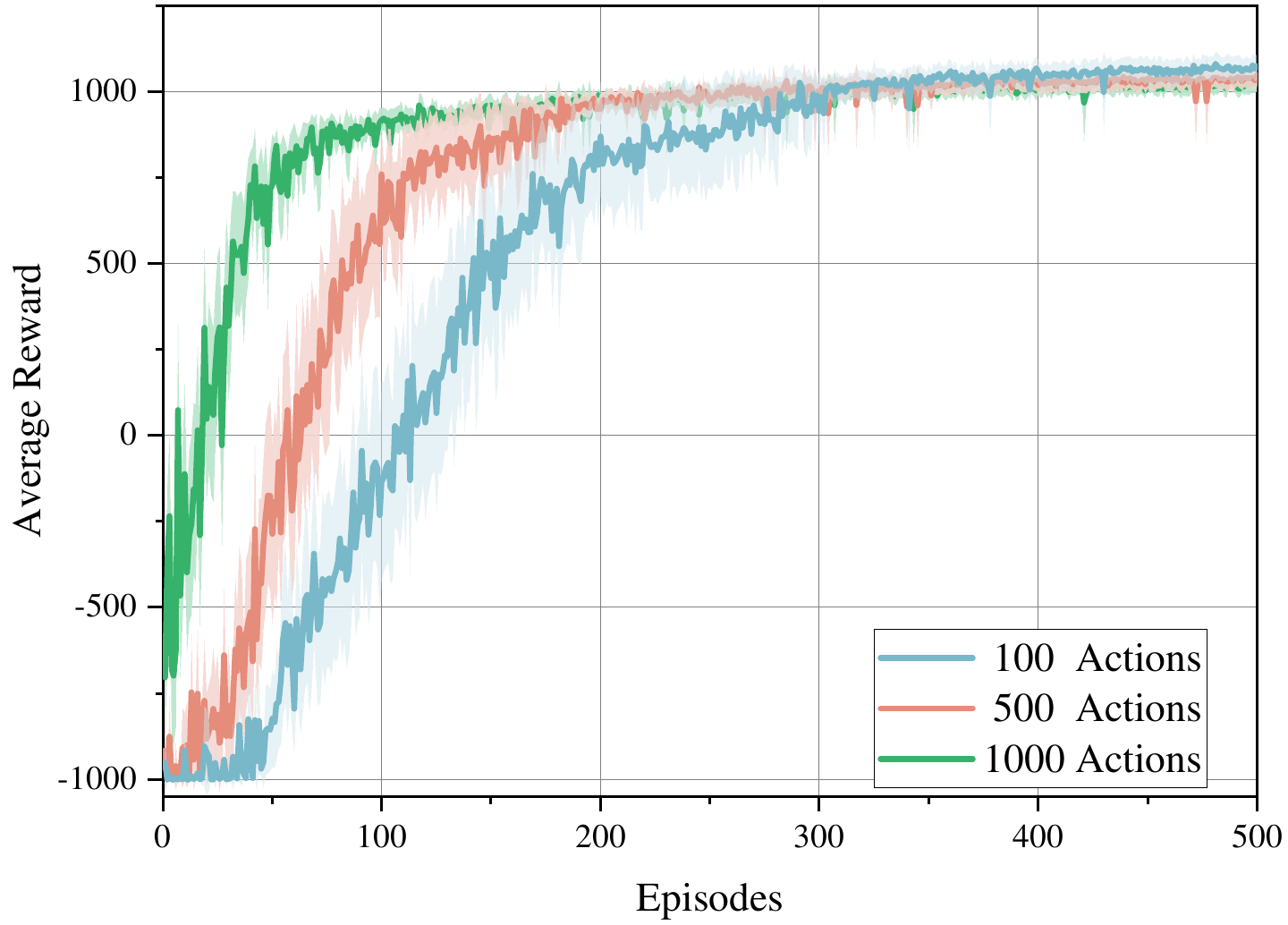}
	\caption{Learning curves of the agent with varying action space sizes in all original training environments over three independent runs. Each curve is the average reward over three independent runs, while the shaded area denotes the 95\% confidence intervals.}
	\label{fig: exp1-learning_curve}
\end{figure}

\begin{figure}[htbp]
	\centering
    \includegraphics[width=0.35\textwidth]{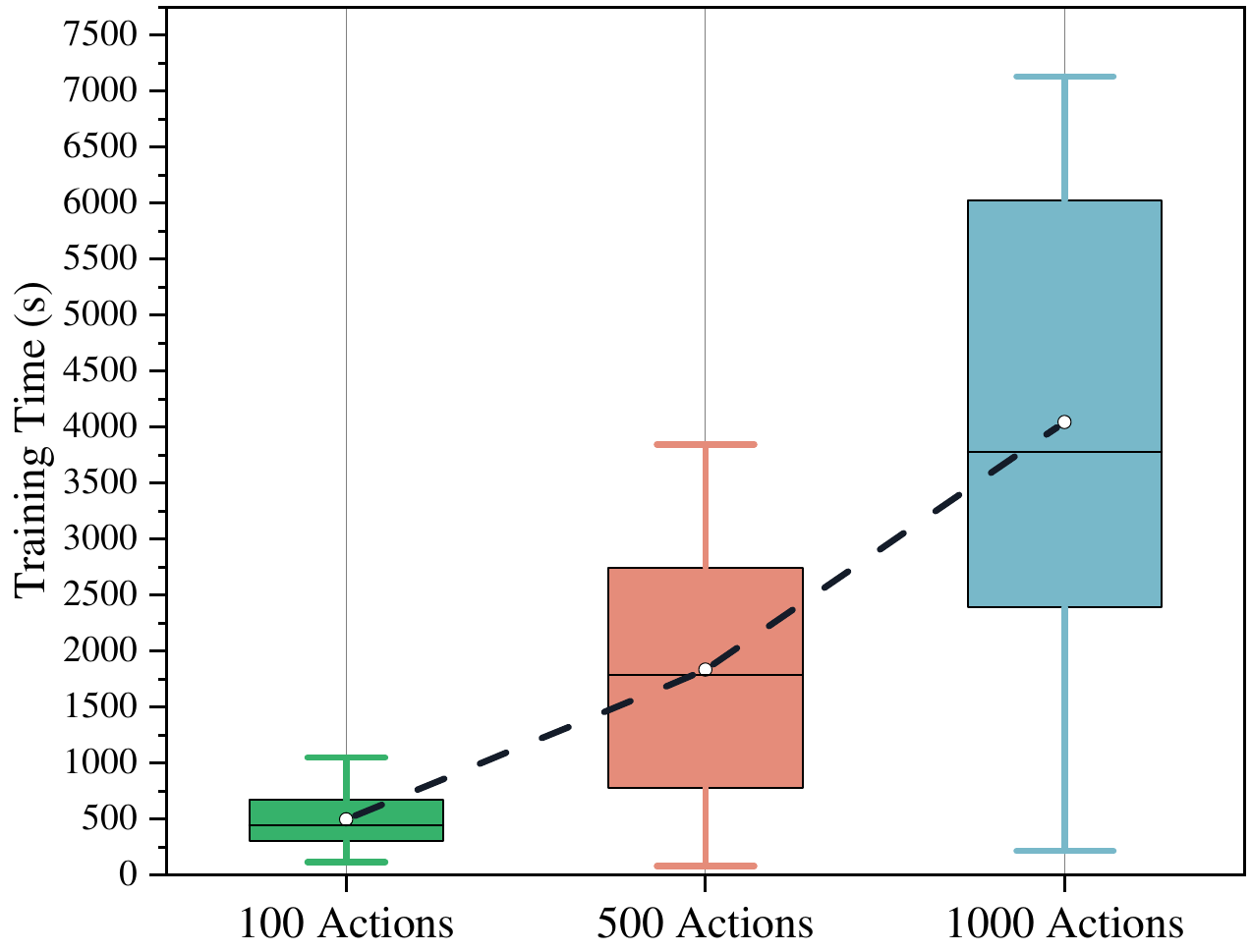}
	\caption{Training time of the agent with varying action space sizes in all original training environments. The box plot summarizes the distribution of training time across 500 episodes for each action space. The dashed line connects the average training time across different action spaces.}
	\label{fig: exp1-training_time}
\end{figure}

Being able to learn in various real environments is the foundation skill for pentesting agents. 
In this part, we train the agent to learn policies from scratch in all original training environments that are unknown to the agent.  We compare the learning performance of the agent across different action spaces ($
\left | \mathcal{A}  \right |  \in {\left \{ 100,500,1000 \right \} }$). 
Fig.\ref{fig: exp1-learning_curve} and Fig.\ref{fig: exp1-training_time} display the agent's learning curve and training time, respectively.

The experimental results demonstrate that the agent can learn policy in training environments. 
However, as the action space increases, there is a decrease in learning efficiency, a slower convergence rate of the learning curve, and an increase in training time accordingly.  
This phenomenon arises because a larger action space requires more extensive exploration to identify effective actions, thereby prolonging the time needed for the agent to discover the optimal policy. Additionally, optimizing the policy model becomes more computationally demanding with increased iterations and computational effort as the action space expands.  

As more vulnerabilities are disclosed, the agent inevitably requires a larger action space to address the expanding attack surface. Consequently, enhancing the learning efficiency of the agent in larger action spaces becomes an interesting research topic. This paper proposes a solution by enhancing the agent's generalization ability. This enables the agent to draw inferences across environments, achieving zero-shot generalization in similar settings and rapid policy adaptation in dissimilar ones. The experimental results supporting these claims are presented in the following sections. 

During interaction with the original training environments, the agent can gather the host configuration data (e.g., ports, services, web fingerprint, etc.) and save them in JSON format as the simulated copy of the real environment.  
This approach not only provides realistic simulated environments for agent training, effectively addressing training environment dilemma, but also enables the synthesis of additional diverse simulation environments using large language models, thereby augmenting the environment variety. 
An example of this process is shown in Fig.\ref{fig:llm}, and more details of the simulated environments are available in our code repository. 

\subsection{Zero-shot Policy Transfer in Similar Environments (RQ2)}\label{exp2}


In this part, we demonstrate the agent's zero-shot generalization ability in testing environments that have the same vulnerability as the original training environment. 

In particular, the agent is pre-trained using various methods in the original training environment and subsequently transferred its policies to its training environment and testing environments. 
We compare the zero-shot generalization performance  between GAP and two baseline methods: PPO \cite{YANG:193309} and April \cite{AprilAE}. Both baselines are state-of-the-art methods that have been wildly used in autonomous pentesting.

Additionally, we also analyze the effect of the number of meta-training environments on the zero-shot generalization performance of GAP. We use $n$-GAP to denote the use of $n$ meta-training environments in the GAP, where $n \in \left \{ 3,5,8 \right \} $ in our experiments. 

\begin{figure}[htbp]
	\centering
	\includegraphics[width=0.4\textwidth]{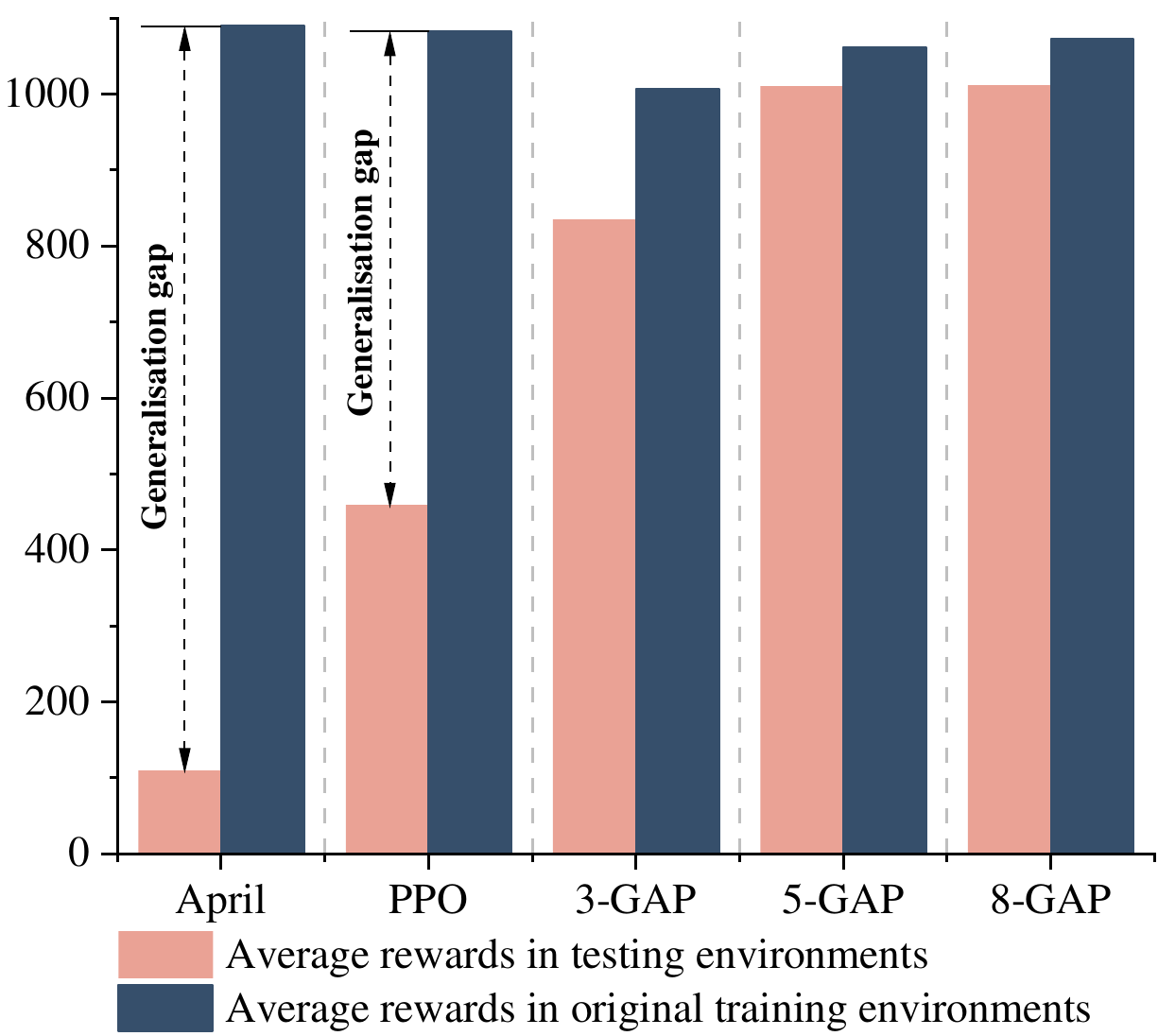}
	\caption{Zero-shot generalization performance in original training environments and testing environments.}
	\label{fig: exp2}
\end{figure}
As shown in Fig.\ref{fig: exp2}, when the agent's policy is transferred to the original training environments, all methods maintain satisfactory performance. However, when the policy is transferred to the testing environments, both baselines experience a noticeable performance loss. This performance gap between the training and testing environments is referred to as the generalization gap. 
Table \ref{tab: generalization_gap} presents the generalization gap (GenGap) of different methods, along with their success rate in the testing environments.
From the results we can see that our proposed framework, GAP, significantly enhances success rates in testing environments compared to the baseline method and bridges the generalization gap, demonstrating improved zero-shot generalization ability.  

\begin{table}[htb]
\centering
\caption{Generalization gap and success rate in testing environments.}
\label{tab: generalization_gap}
\begin{tabular}{>{\centering\arraybackslash}p{0.2\linewidth}>{\centering\arraybackslash}p{0.2\linewidth}>{\centering\arraybackslash}p{0.3\linewidth}}
\hline
\textbf{Method} & \textbf{GenGap} & \textbf{Success Rate} \\ \hline
April           & 990.21                     & 0.14           \\
PPO             & 624.33                    & 0.44           \\
3-GAP           & 171.68& 0.83                \\
5-GAP           & 51.61                  & 0.92           \\
8-GAP           & 61.11                   & 0.91           \\ \hline
\end{tabular}

\end{table}
These results indicate that the use of domain randomization increased both the quantity and diversity of training environments. This exposure enabled the agent to adapt to a wider range of environmental variations during training. Through meta-RL, the agent learned how to extract generalizable policies and biases from the diverse meta-training environments, thereby allowing the agent to effectively generalize its learned policies to new, similar testing environments. 

Additionally, fewer meta-training environments affect the agent's zero-shot generalization ability to some extent. 
However, increasing the number of meta-training environments up to a certain point does not significantly improve generalization performance. This could be due to the fact that the agent reaches a saturation point where additional environment variations do not significantly contribute to further learning.  
Specifically, our experiments reveal that the agent can achieve satisfactory zero-shot generalization performance when trained with five meta-training environments, which serves as the default setting in the subsequent experiments.

\subsection{Few-shot Policy Adaptation in Dissimilar Environments (RQ3)}\label{exp3}

In this part, we evaluate the agent's policy adaptation ability when faced with testing environments that are dissimilar to the original training environments.  
The policy adaptation ability refers to the agent's ability to quickly adapt its learned policy  to unseen environments through policy transfer.  

We use the following four methods to train the agent with action space $\left | A \right | =1000$:
\begin{itemize}
	\item Learn from scratch.  Similar to Section \ref{exp1}, the agent is directly trained using PPO in testing environments without policy transfer.
    \item PPO-Transfer. We pre-train the agent using PPO in original training environments and then transfer its learned policy to testing environments.
	\item April-Transfer. We pre-train the agent using April in  original training environments and then transfer its learned policy to testing environments. April is the state-of-the-art autonomous pentesting framework that has been proven to exhibit strong transfer learning performance \cite{AprilAE}. 	
    \item  GAP-Transfer. Following the Real-to-Sim-to-Real pipeline, the agent is pre-trained in the original training environment, and then its generalization is improved by adopting domain randomization and meta-RL training. Finally, we transfer its learned policy to testing environments. 
\end{itemize}

\begin{figure}[htb]
	\centering
	\includegraphics[width=0.4\textwidth]{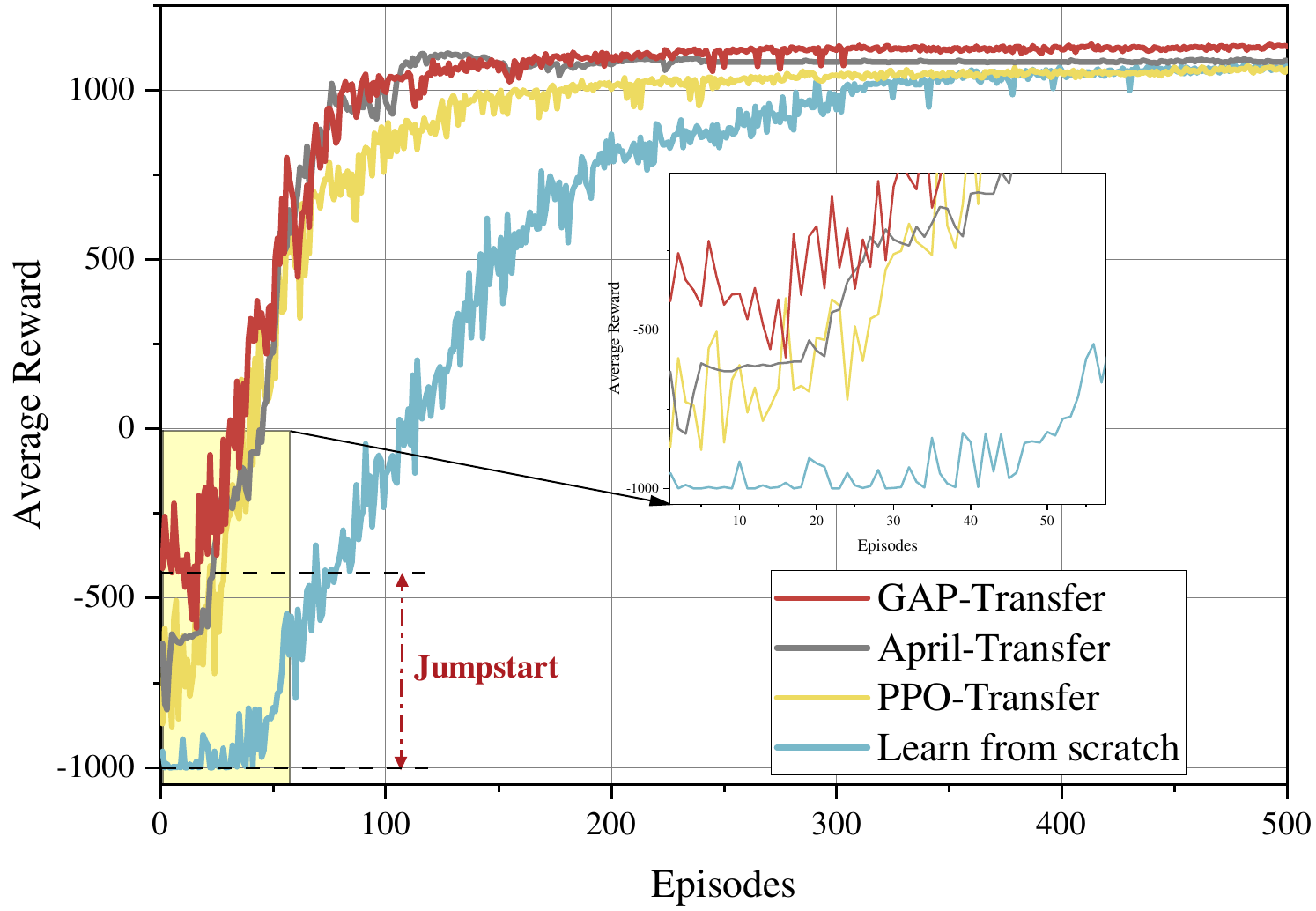}
	\caption{Learning curves of different methods  in testing environments. }
	\label{fig: exp3-learning_curve}
\end{figure}

\begin{table}[htpb]
\centering
\caption{Jumpstart performance of different methods.}
\label{tab:jump}
\begin{tabular}{lc}
\hline
\textbf{Method}        & \multicolumn{1}{l}{\textbf{Jumpstart}} \\ \hline
GAP-Transfer                    & 589.22                                   \\
April-Transfer         & 367.21                                   \\
PPO-Transfer           & 129.83                                   \\
Learn from scratch& 47.33                                    \\ \hline
                       & \multicolumn{1}{l}{}                   
\end{tabular}

\end{table}

\begin{figure}[htb]
	\centering
	\includegraphics[width=0.3\textwidth]{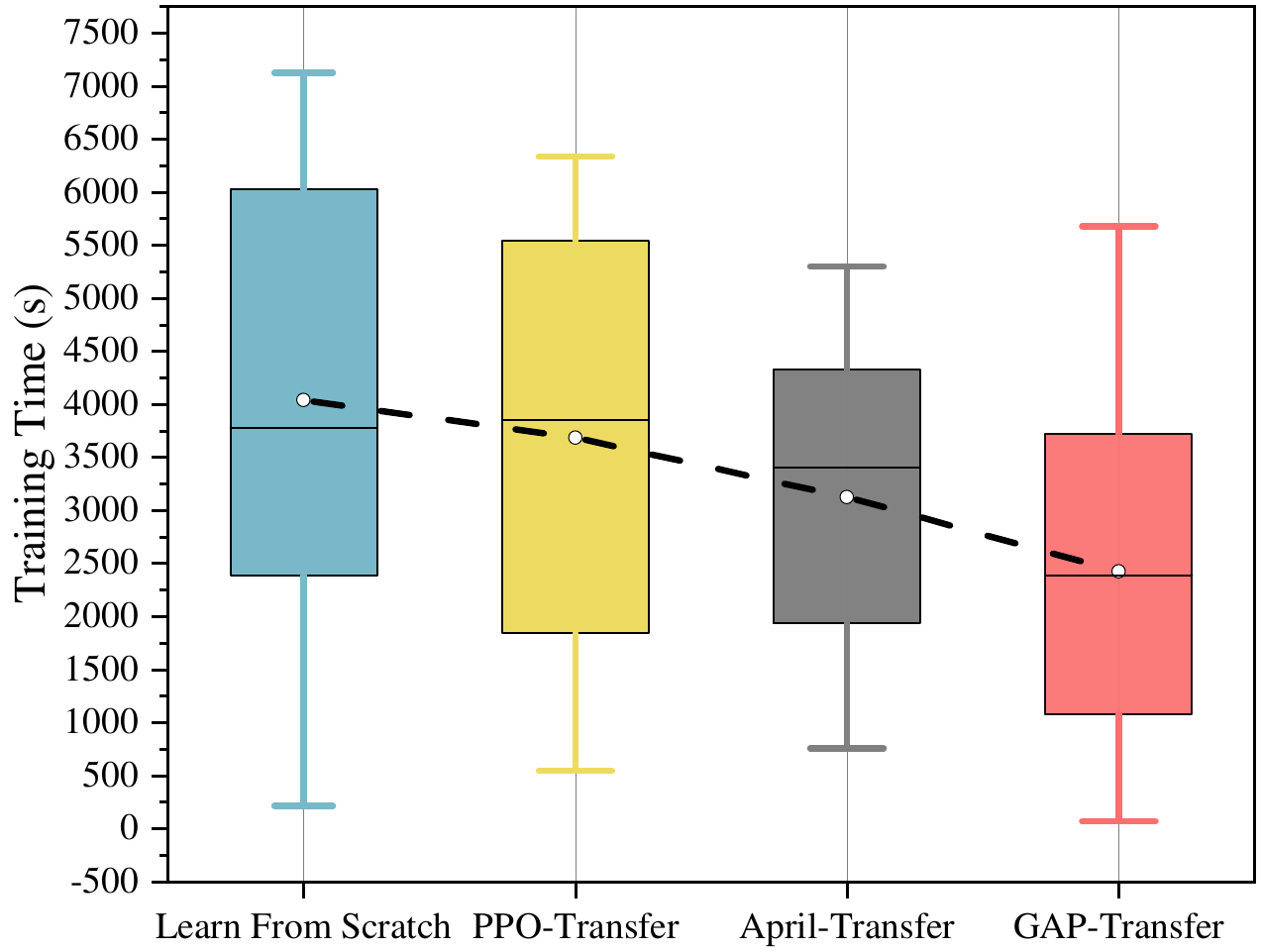}
	\caption{Training time of the agent in testing environments over three independent runs. The box plot summarizes the distribution of training time across 500 episodes, while the dashed line connects the average value.}
	\label{fig: exp3-training_time}
\end{figure}

Fig.\ref{fig: exp3-learning_curve} and Fig.\ref{fig: exp3-training_time} display the agent's learning curve and training time in testing environments, respectively. 
The results show that, compared to learning from scratch, the other three methods enhance convergence speed and reduce training time through policy transfer. Among them, GAP stands out as the most effective, reducing average training time by approximately 40\% (from 4045 seconds to 2429 seconds). 

\begin{figure}[htb]
	\centering
	\includegraphics[width=0.3\textwidth]{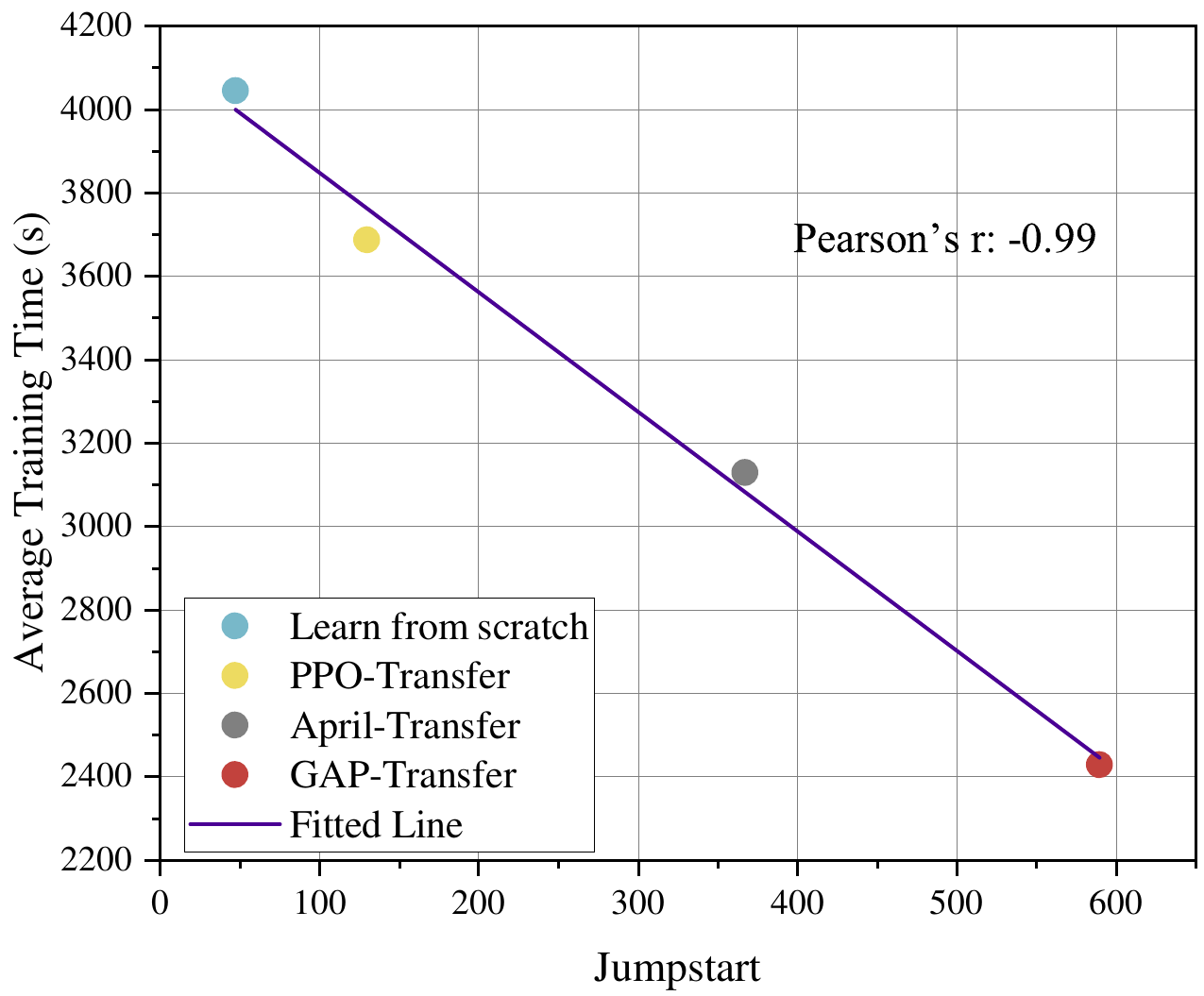}
	\caption{Correlation between jumpstart and average training time of different methods.}
	\label{fig:jump-time}
\end{figure}

Furthermore, as shown in Fig.\ref{fig: exp3-learning_curve} and Fig.\ref{fig: exp3-training_time} , it can be observed that while GAP-Transfer and April-Transfer exhibit similar convergence speeds, GAP-Transfer reduces training time by approximately 22\% compared to April-Transfer (from 3129 seconds to 2429 seconds). This reduction is related to the jumpstart performance, as evidenced by Table \ref{tab:jump} and  Fig.\ref{fig:jump-time}. 

Table \ref{tab:jump} presents the jumpstart performance of different methods, from which we can see  that GAP-Transfer demonstrates superior jumpstart performance than others.  
Besides, Fig.\ref{fig:jump-time} illustrates a strong negative linear correlation between jumpstart performance and average training time, with a Pearson correlation coefficient (Pearson's r) close to $-0.99$. 
A better jumpstart allows the agent to explore effective actions more quickly during initial training in new environments, indicating that the agent can generalize knowledge from past experiences and apply it effectively to new situations. The outstanding performance of GAP-Transfer can be attributed to the incorporation of a meta-learning mechanism, which empowers the agent to improve its policy adaptation ability by learning how to learn. 




\section{Discussion}

This section aims to provide a critical discussion of the limitations of our work and potential avenues for future improvement.

Firstly, our work aims to enhance the policy generalization ability of pentesting agents, which is essential for their widespread applicability.  And we also provide a feasible solution for training pentesting agents in actual machines.
Given the constraints of real-world environment safety and the availability of exploit code, we validate the application potential of these agents in real environments by testing them in a limited number of virtual machine environments. These environments predominantly feature vulnerabilities that enable remote command execution. 
Admittedly, our vulnerability environments are somewhat idealized and overlook the impacts that complex real-world environments may have on pentesting agents, such as effects from network protocols or honeypot environments. Effective autonomous pentesting by agents in complex environments hinges on more precise environmental observation capabilities and a rich, fine-grained action space. This will be a focal point of our future research endeavors. 

Besides, GAP aims to train the agent to realize drawing inferences from one to another. However, another intriguing area that demands exploration is achieving "lifelong learning" in agents. Lifelong learning focuses on enabling agents to avoid forgetting past tasks when faced with numerous new tasks. GAP provides a foundational framework for realizing this capability, which we intend to explore further in our future research. 


Furthermore, in GAP, we utilize a general LLM for generating synthetic environments, which are evaluated for validity by human experts. In the future, we will explore methods for generating pentesting simulation environments using domain-specific LLMs, thereby improving the quality of these environments.

Finally, in our work, the reward function's design relies on experts' subjective estimates. However, assessing the reward function presents a challenge since there's no objective standard defining an "optimal" strategy \cite{Lore}. One potential avenue for future research is to explore automated methods (e.g., inverse reinforcement learning \cite{Metelli24}) for generating reward functions. 

\section{Conclusion}

In this paper, we present GAP, an autonomous pentesting framework for efficient policy training in realistic environments and for training generalizable agents capable of drawing inferences about other cases from one instance -- a key to the broad application of autonomous pentesting agents. 
To achieve this, GAP introduces a Real-to-Sim-to-Real pipeline that (a) enables end-to-end policy learning in unknown realistic environments while constructing realistic simulation analogs; (b) improves agents' generalization ability by leveraging domain randomization and meta-RL learning.
The preliminary evaluations demonstrate that GAP allows pentesting agents for end-to-end policy learning in realistic environments, bridging the generalization gap for zero-shot policy transfer in similar environments, and facilitating rapid policy adaptation in dissimilar environments. 
GAP fills a critical research gap as the first framework of its kind. 
By introducing GAP, we not only contribute to the field of autonomous pentesting but also provide a feasible framework that may be used in other applications.

\bibliographystyle{elsarticle-num} 
\bibliography{myrefs}

\begin{thebibliography}{10}
\expandafter\ifx\csname url\endcsname\relax
  \def\url#1{\texttt{#1}}\fi
\expandafter\ifx\csname urlprefix\endcsname\relax\def\urlprefix{URL }\fi
\expandafter\ifx\csname href\endcsname\relax
  \def\href#1#2{#2} \def\path#1{#1}\fi

\bibitem{Lore}
H.~Holm, Lore a red team emulation tool, {IEEE} Trans. Dependable Secur. Comput. 20~(2) (2023) 1596--1608.
\newblock \href {https://doi.org/10.1109/TDSC.2022.3160792} {\path{doi:10.1109/TDSC.2022.3160792}}.

\bibitem{ChenHJLW22}
X.~Chen, J.~Hu, C.~Jin, L.~Li, L.~Wang, Understanding domain randomization for sim-to-real transfer, in: The Tenth International Conference on Learning Representations, {ICLR} 2022, Virtual Event, April 25-29, 2022, OpenReview.net, 2022.

\bibitem{schwartz2019autonomous}
J.~Schwartz, H.~Kurniawati, Autonomous penetration testing using reinforcement learning, arXiv preprint arXiv:1905.05965 (2019).

\bibitem{zennaro2020modeling}
F.~M. Zennaro, L.~Erdodi, Modeling penetration testing with reinforcement learning using capture-the-flag challenges and tabular q-learning, arXiv preprint arXiv:2005.12632 (2020).

\bibitem{Hu2020AutomatedPT}
Z.~Hu, R.~Beuran, Y.~Tan, Automated penetration testing using deep reinforcement learning, 2020 IEEE European Symposium on Security and Privacy Workshops (EuroS\&PW) (2020) 2--10.

\bibitem{Tran2021DeepHR}
K.~Tran, A.~Akella, M.~Standen, J.~Kim, D.~Bowman, T.~J. Richer, C.-T. L.~I. One, I.~Two, Deep hierarchical reinforcement agents for automated penetration testing, ArXiv abs/2109.06449 (2021).

\bibitem{abs-2202-10630}
Y.~Yang, X.~Liu, Behaviour-diverse automatic penetration testing: {A} curiosity-driven multi-objective deep reinforcement learning approach, CoRR abs/2202.10630 (2022).
\newblock \href {http://arxiv.org/abs/2202.10630} {\path{arXiv:2202.10630}}.

\bibitem{Chen2023GAILPTAI}
J.~Chen, S.~Hu, H.~Zheng, C.~Xing, G.~Zhang, Gail-pt: An intelligent penetration testing framework with generative adversarial imitation learning, Comput. Secur. 126 (2023) 103055.

\bibitem{YangCFL23}
Y.~Yang, M.~Chen, H.~Fu, X.~Liu, Settron: Towards better generalisation in penetration testing with reinforcement learning, in: {IEEE} Global Communications Conference, {GLOBECOM} 2023, Kuala Lumpur, Malaysia, December 4-8, 2023, 2023, pp. 4662--4667.
\newblock \href {https://doi.org/10.1109/GLOBECOM54140.2023.10437804} {\path{doi:10.1109/GLOBECOM54140.2023.10437804}}.

\bibitem{10681147}
Q.~Li, R.~Wang, D.~Li, F.~Shi, M.~Zhang, A.~Chattopadhyay, Y.~Shen, Y.~Li, Dynpen: Automated penetration testing in dynamic network scenarios using deep reinforcement learning, IEEE Transactions on Information Forensics and Security 19 (2024) 8966--8981.
\newblock \href {https://doi.org/10.1109/TIFS.2024.3461950} {\path{doi:10.1109/TIFS.2024.3461950}}.

\bibitem{schwartz2019nasim}
J.~Schwartz, H.~Kurniawatti, \href{https://networkattacksimulator.readthedocs.io/}{Nasim: Network attack simulator} (2019).
\newline\urlprefix\url{https://networkattacksimulator.readthedocs.io/}

\bibitem{msft:cyberbattlesim}
M.~D.~R. Team., \href{https://github.com/microsoft/cyberbattlesim}{Cyberbattlesim}, created by Christian Seifert, Michael Betser, William Blum, James Bono, Kate Farris, Emily Goren, Justin Grana, Kristian Holsheimer, Brandon Marken, Joshua Neil, Nicole Nichols, Jugal Parikh, Haoran Wei. (2021).
\newline\urlprefix\url{https://github.com/microsoft/cyberbattlesim}

\bibitem{AprilAE}
S.~Zhou, J.~Liu, Y.~Lu, J.~Yang, D.~Hou, Y.~Zhang, S.~Hu, April: towards scalable and transferable autonomous penetration testing in large action space via action embedding, IEEE Transactions on Dependable and Secure Computing (2024) 1--17\href {https://doi.org/10.1109/TDSC.2024.3518500} {\path{doi:10.1109/TDSC.2024.3518500}}.

\bibitem{WangKSF20}
K.~Wang, B.~Kang, J.~Shao, J.~Feng, Improving generalization in reinforcement learning with mixture regularization, in: H.~Larochelle, M.~Ranzato, R.~Hadsell, M.~Balcan, H.~Lin (Eds.), Advances in Neural Information Processing Systems 33: Annual Conference on Neural Information Processing Systems 2020, NeurIPS 2020, December 6-12, 2020, virtual, 2020.

\bibitem{SongJTDN20}
X.~Song, Y.~Jiang, S.~Tu, Y.~Du, B.~Neyshabur, Observational overfitting in reinforcement learning, in: 8th International Conference on Learning Representations, {ICLR} 2020, Addis Ababa, Ethiopia, April 26-30, 2020, OpenReview.net, 2020.

\bibitem{KirkZGR23}
R.~Kirk, A.~Zhang, E.~Grefenstette, T.~Rockt{\"{a}}schel, A survey of zero-shot generalisation in deep reinforcement learning, J. Artif. Intell. Res. 76 (2023) 201--264.
\newblock \href {https://doi.org/10.1613/JAIR.1.14174} {\path{doi:10.1613/JAIR.1.14174}}.

\bibitem{CobbeKHKS19}
K.~Cobbe, O.~Klimov, C.~Hesse, T.~Kim, J.~Schulman, Quantifying generalization in reinforcement learning, in: K.~Chaudhuri, R.~Salakhutdinov (Eds.), Proceedings of the 36th International Conference on Machine Learning, {ICML} 2019, 9-15 June 2019, Long Beach, California, {USA}, Vol.~97 of Proceedings of Machine Learning Research, {PMLR}, 2019, pp. 1282--1289.

\bibitem{TobinFRSZA17}
J.~Tobin, R.~Fong, A.~Ray, J.~Schneider, W.~Zaremba, P.~Abbeel, Domain randomization for transferring deep neural networks from simulation to the real world, in: 2017 {IEEE/RSJ} International Conference on Intelligent Robots and Systems, {IROS} 2017, Vancouver, BC, Canada, September 24-28, 2017, {IEEE}, 2017, pp. 23--30.
\newblock \href {https://doi.org/10.1109/IROS.2017.8202133} {\path{doi:10.1109/IROS.2017.8202133}}.

\bibitem{HorvathEIHF23}
D.~Horv{\'{a}}th, G.~Erd{\"{o}}s, Z.~Istenes, T.~Horv{\'{a}}th, S.~F{\"{o}}ldi, Object detection using sim2real domain randomization for robotic applications, {IEEE} Trans. Robotics 39~(2) (2023) 1225--1243.
\newblock \href {https://doi.org/10.1109/TRO.2022.3207619} {\path{doi:10.1109/TRO.2022.3207619}}.

\bibitem{LiZL023}
Z.~Li, H.~Zhu, Z.~Lu, M.~Yin, Synthetic data generation with large language models for text classification: Potential and limitations, in: H.~Bouamor, J.~Pino, K.~Bali (Eds.), Proceedings of the 2023 Conference on Empirical Methods in Natural Language Processing, {EMNLP} 2023, Singapore, December 6-10, 2023, Association for Computational Linguistics, 2023, pp. 10443--10461.
\newblock \href {https://doi.org/10.18653/V1/2023.EMNLP-MAIN.647} {\path{doi:10.18653/V1/2023.EMNLP-MAIN.647}}.

\bibitem{abs-2403-04190}
X.~Guo, Y.~Chen, Generative {AI} for synthetic data generation: Methods, challenges and the future, CoRR abs/2403.04190 (2024).
\newblock \href {http://arxiv.org/abs/2403.04190} {\path{arXiv:2403.04190}}, \href {https://doi.org/10.48550/ARXIV.2403.04190} {\path{doi:10.48550/ARXIV.2403.04190}}.

\bibitem{HospedalesAMS22}
T.~M. Hospedales, A.~Antoniou, P.~Micaelli, A.~J. Storkey, Meta-learning in neural networks: {A} survey, {IEEE} Trans. Pattern Anal. Mach. Intell. 44~(9) (2022) 5149--5169.
\newblock \href {https://doi.org/10.1109/TPAMI.2021.3079209} {\path{doi:10.1109/TPAMI.2021.3079209}}.

\bibitem{YeZGZ24}
D.~Ye, T.~Zhu, K.~Gao, W.~Zhou, Defending against label-only attacks via meta-reinforcement learning, {IEEE} Trans. Inf. Forensics Secur. 19 (2024) 3295--3308.
\newblock \href {https://doi.org/10.1109/TIFS.2024.3357292} {\path{doi:10.1109/TIFS.2024.3357292}}.

\bibitem{abs-2301-08028}
J.~Beck, R.~Vuorio, E.~Z. Liu, Z.~Xiong, L.~M. Zintgraf, C.~Finn, S.~Whiteson, A survey of meta-reinforcement learning, CoRR abs/2301.08028 (2023).
\newblock \href {http://arxiv.org/abs/2301.08028} {\path{arXiv:2301.08028}}, \href {https://doi.org/10.48550/ARXIV.2301.08028} {\path{doi:10.48550/ARXIV.2301.08028}}.

\bibitem{FinnAL17}
C.~Finn, P.~Abbeel, S.~Levine, Model-agnostic meta-learning for fast adaptation of deep networks, in: D.~Precup, Y.~W. Teh (Eds.), Proceedings of the 34th International Conference on Machine Learning, {ICML} 2017, Sydney, NSW, Australia, 6-11 August 2017, Vol.~70 of Proceedings of Machine Learning Research, {PMLR}, 2017, pp. 1126--1135.

\bibitem{Botvinick2019ReinforcementLF}
M.~M. Botvinick, S.~Ritter, J.~X. Wang, Z.~Kurth-Nelson, C.~Blundell, D.~Hassabis, Reinforcement learning, fast and slow, Trends in Cognitive Sciences 23 (2019) 408--422.

\bibitem{schneier1999attack}
B.~Schneier, Attack trees, Dr. Dobb’s journal 24~(12) (1999) 21--29.

\bibitem{Sheyner2002AutomatedGA}
O.~Sheyner, J.~W. Haines, S.~Jha, R.~Lippmann, J.~M. Wing, Automated generation and analysis of attack graphs, in: 2002 {IEEE} Symposium on Security and Privacy, Berkeley, California, USA, May 12-15, 2002, {IEEE} Computer Society, 2002, pp. 273--284.
\newblock \href {https://doi.org/10.1109/SECPRI.2002.1004377} {\path{doi:10.1109/SECPRI.2002.1004377}}.

\bibitem{JmalHBIKW24}
H.~Jmal, F.~B. Hmida, N.~Basta, M.~Ikram, M.~A. K{\^{a}}afar, A.~Walker, {SPGNN-API:} {A} transferable graph neural network for attack paths identification and autonomous mitigation, {IEEE} Trans. Inf. Forensics Secur. 19 (2024) 1601--1613.
\newblock \href {https://doi.org/10.1109/TIFS.2023.3338965} {\path{doi:10.1109/TIFS.2023.3338965}}.

\bibitem{Vinyals2019GrandmasterLI}
O.~Vinyals, I.~Babuschkin, W.~M. Czarnecki, M.~Mathieu, A.~Dudzik, J.~Chung, D.~H. Choi, R.~Powell, T.~Ewalds, P.~Georgiev, J.~Oh, D.~Horgan, M.~Kroiss, I.~Danihelka, A.~Huang, L.~Sifre, T.~Cai, J.~P. Agapiou, M.~Jaderberg, A.~S. Vezhnevets, R.~Leblond, T.~Pohlen, V.~Dalibard, D.~Budden, Y.~Sulsky, J.~Molloy, T.~L. Paine, C.~Gulcehre, Z.~Wang, T.~Pfaff, Y.~Wu, R.~Ring, D.~Yogatama, D.~W{\"u}nsch, K.~McKinney, O.~Smith, T.~Schaul, T.~P. Lillicrap, K.~Kavukcuoglu, D.~Hassabis, C.~Apps, D.~Silver, Grandmaster level in starcraft ii using multi-agent reinforcement learning, Nature (2019) 1--5.

\bibitem{Feng2023DenseRL}
S.~Feng, H.~Sun, X.~Yan, H.~Zhu, Z.~Zou, S.~Shen, H.~X. Liu, Dense reinforcement learning for safety validation of autonomous vehicles, Nature 615 (2023) 620 -- 627.

\bibitem{BoZZHLZL24}
L.~Bo, T.~Zhang, H.~Zhang, J.~Hong, M.~Liu, C.~Zhang, B.~Liu, 3d {UAV} path planning in unknown environment: {A} transfer reinforcement learning method based on low-rank adaption, Adv. Eng. Informatics 62 (2024) 102920.
\newblock \href {https://doi.org/10.1016/J.AEI.2024.102920} {\path{doi:10.1016/J.AEI.2024.102920}}.

\bibitem{LiLYP22}
L.~Li, Y.~Luo, J.~Yang, L.~Pu, Reinforcement learning enabled intelligent energy attack in green iot networks, {IEEE} Trans. Inf. Forensics Secur. 17 (2022) 644--658.
\newblock \href {https://doi.org/10.1109/TIFS.2022.3149148} {\path{doi:10.1109/TIFS.2022.3149148}}.

\bibitem{IlicDVMP24}
N.~Ilic, D.~Dasic, M.~Vucetic, A.~Makarov, R.~Petrovic, Distributed web hacking by adaptive consensus-based reinforcement learning, Artif. Intell. 326 (2024) 104032.
\newblock \href {https://doi.org/10.1016/J.ARTINT.2023.104032} {\path{doi:10.1016/J.ARTINT.2023.104032}}.

\bibitem{YANG:193309}
Y.~YANG, L.~CHEN, S.~LIU, L.~WANG, H.~FU, X.~LIU, Z.~CHEN, Behaviour-diverse automatic penetration testing: a coverage-based deep reinforcement learning approach, Frontiers of Computer Science 19~(3) (2025) 193309.
\newblock \href {https://doi.org/10.1007/s11704-024-3380-1} {\path{doi:10.1007/s11704-024-3380-1}}.

\bibitem{de}
I.~Takaesu, \href{https://www.mbsd.jp/blog/20180228.html}{Deepexploit} (2018).
\newline\urlprefix\url{https://www.mbsd.jp/blog/20180228.html}

\bibitem{Maeda2021AutomatingPW}
R.~Maeda, M.~Mimura, Automating post-exploitation with deep reinforcement learning, Comput. Secur. 100 (2021) 102108.

\bibitem{8575297}
J.~Tremblay, A.~Prakash, D.~Acuna, M.~Brophy, V.~Jampani, C.~Anil, T.~To, E.~Cameracci, S.~Boochoon, S.~Birchfield, Training deep networks with synthetic data: Bridging the reality gap by domain randomization, in: 2018 IEEE/CVF Conference on Computer Vision and Pattern Recognition Workshops (CVPRW), 2018, pp. 1082--10828.
\newblock \href {https://doi.org/10.1109/CVPRW.2018.00143} {\path{doi:10.1109/CVPRW.2018.00143}}.

\bibitem{PengAZA18}
X.~B. Peng, M.~Andrychowicz, W.~Zaremba, P.~Abbeel, Sim-to-real transfer of robotic control with dynamics randomization, in: 2018 {IEEE} International Conference on Robotics and Automation, {ICRA} 2018, Brisbane, Australia, May 21-25, 2018, {IEEE}, 2018, pp. 1--8.
\newblock \href {https://doi.org/10.1109/ICRA.2018.8460528} {\path{doi:10.1109/ICRA.2018.8460528}}.

\bibitem{TiboniPTA23}
G.~Tiboni, A.~Protopapa, T.~Tommasi, G.~Averta, \href{https://doi.org/10.1109/IROS55552.2023.10342537}{Domain randomization for robust, affordable and effective closed-loop control of soft robots}, in: {IROS}, 2023, pp. 612--619.
\newblock \href {https://doi.org/10.1109/IROS55552.2023.10342537} {\path{doi:10.1109/IROS55552.2023.10342537}}.
\newline\urlprefix\url{https://doi.org/10.1109/IROS55552.2023.10342537}

\bibitem{ChenTRBY23}
J.~Chen, D.~Tam, C.~Raffel, M.~Bansal, D.~Yang, An empirical survey of data augmentation for limited data learning in {NLP}, Trans. Assoc. Comput. Linguistics 11 (2023) 191--211.
\newblock \href {https://doi.org/10.1162/TACL\_A\_00542} {\path{doi:10.1162/TACL\_A\_00542}}.

\bibitem{long2024llmsdrivensyntheticdatageneration}
L.~{Long}, R.~{Wang}, R.~{Xiao}, J.~{Zhao}, X.~{Ding}, G.~{Chen}, H.~{Wang}, {On LLMs-Driven Synthetic Data Generation, Curation, and Evaluation: A Survey}, arXiv e-prints (2024) arXiv:2406.15126\href {http://arxiv.org/abs/2406.15126} {\path{arXiv:2406.15126}}, \href {https://doi.org/10.48550/arXiv.2406.15126} {\path{doi:10.48550/arXiv.2406.15126}}.

\bibitem{abs-2402-14568}
J.~Ye, N.~Xu, Y.~Wang, J.~Zhou, Q.~Zhang, T.~Gui, X.~Huang, {LLM-DA:} data augmentation via large language models for few-shot named entity recognition, CoRR abs/2402.14568 (2024).
\newblock \href {http://arxiv.org/abs/2402.14568} {\path{arXiv:2402.14568}}, \href {https://doi.org/10.48550/ARXIV.2402.14568} {\path{doi:10.48550/ARXIV.2402.14568}}.

\bibitem{abs-2403-02990}
B.~Ding, C.~Qin, R.~Zhao, T.~Luo, X.~Li, G.~Chen, W.~Xia, J.~Hu, A.~T. Luu, S.~Joty, Data augmentation using llms: Data perspectives, learning paradigms and challenges, CoRR abs/2403.02990 (2024).
\newblock \href {http://arxiv.org/abs/2403.02990} {\path{arXiv:2403.02990}}, \href {https://doi.org/10.48550/ARXIV.2403.02990} {\path{doi:10.48550/ARXIV.2403.02990}}.

\bibitem{abs-2305-14288}
C.~Whitehouse, M.~Choudhury, A.~F. Aji, Llm-powered data augmentation for enhanced crosslingual performance, CoRR abs/2305.14288 (2023).
\newblock \href {http://arxiv.org/abs/2305.14288} {\path{arXiv:2305.14288}}, \href {https://doi.org/10.48550/ARXIV.2305.14288} {\path{doi:10.48550/ARXIV.2305.14288}}.

\bibitem{JuJGNL22}
H.~Ju, R.~Juan, R.~Gomez, K.~Nakamura, G.~Li, Transferring policy of deep reinforcement learning from simulation to reality for robotics, Nat. Mac. Intell. 4~(12) (2022) 1077--1087.
\newblock \href {https://doi.org/10.1038/S42256-022-00573-6} {\path{doi:10.1038/S42256-022-00573-6}}.

\bibitem{DuanSCBSA16}
Y.~Duan, J.~Schulman, X.~Chen, P.~L. Bartlett, I.~Sutskever, P.~Abbeel, Rl$^2$: Fast reinforcement learning via slow reinforcement learning, CoRR abs/1611.02779 (2016).
\newblock \href {http://arxiv.org/abs/1611.02779} {\path{arXiv:1611.02779}}.

\bibitem{ArndtHGK20}
K.~Arndt, M.~Hazara, A.~Ghadirzadeh, V.~Kyrki, Meta reinforcement learning for sim-to-real domain adaptation, in: 2020 {IEEE} International Conference on Robotics and Automation, {ICRA} 2020, Paris, France, May 31 - August 31, 2020, {IEEE}, 2020, pp. 2725--2731.
\newblock \href {https://doi.org/10.1109/ICRA40945.2020.9196540} {\path{doi:10.1109/ICRA40945.2020.9196540}}.

\bibitem{sutton2018reinforcement}
R.~S. Sutton, A.~G. Barto, Reinforcement learning: An introduction, MIT press, 2018.

\bibitem{BingLHK23}
Z.~Bing, D.~Lerch, K.~Huang, A.~C. Knoll, Meta-reinforcement learning in non-stationary and dynamic environments, {IEEE} Trans. Pattern Anal. Mach. Intell. 45~(3) (2023) 3476--3491.
\newblock \href {https://doi.org/10.1109/TPAMI.2022.3185549} {\path{doi:10.1109/TPAMI.2022.3185549}}.

\bibitem{LyleRDKG22}
C.~Lyle, M.~Rowland, W.~Dabney, M.~Kwiatkowska, Y.~Gal, Learning dynamics and generalization in deep reinforcement learning, in: K.~Chaudhuri, S.~Jegelka, L.~Song, C.~Szepesv{\'{a}}ri, G.~Niu, S.~Sabato (Eds.), International Conference on Machine Learning, {ICML} 2022, 17-23 July 2022, Baltimore, Maryland, {USA}, Vol. 162 of Proceedings of Machine Learning Research, {PMLR}, 2022, pp. 14560--14581.

\bibitem{TaigaAFCB23}
A.~A. Ta{\"{\i}}ga, R.~Agarwal, J.~Farebrother, A.~C. Courville, M.~G. Bellemare, Investigating multi-task pretraining and generalization in reinforcement learning, in: The Eleventh International Conference on Learning Representations, {ICLR} 2023, Kigali, Rwanda, May 1-5, 2023, OpenReview.net, 2023.

\bibitem{ZhaoQW20}
W.~Zhao, J.~P. Queralta, T.~Westerlund, Sim-to-real transfer in deep reinforcement learning for robotics: a survey, in: 2020 {IEEE} Symposium Series on Computational Intelligence, {SSCI} 2020, Canberra, Australia, December 1-4, 2020, {IEEE}, 2020, pp. 737--744.
\newblock \href {https://doi.org/10.1109/SSCI47803.2020.9308468} {\path{doi:10.1109/SSCI47803.2020.9308468}}.

\bibitem{wang-2021-TSDAE}
K.~Wang, N.~Reimers, I.~Gurevych, Tsdae: Using transformer-based sequential denoising auto-encoderfor unsupervised sentence embedding learning, arXiv preprint arXiv:2104.06979 (4 2021).

\bibitem{PPO}
J.~Schulman, F.~Wolski, P.~Dhariwal, A.~Radford, O.~Klimov, Proximal policy optimization algorithms, CoRR abs/1707.06347 (2017).
\newblock \href {http://arxiv.org/abs/1707.06347} {\path{arXiv:1707.06347}}.

\bibitem{glm2024chatglm}
{Team GLM}, {:}, A.~{Zeng}, B.~{Xu}, B.~{Wang}, C.~{Zhang}, D.~{Yin}, D.~{Rojas}, G.~{Feng}, H.~{Zhao}, H.~{Lai}, H.~{Yu}, H.~{Wang}, J.~{Sun}, J.~{Zhang}, J.~{Cheng}, J.~{Gui}, J.~{Tang}, J.~{Zhang}, J.~{Li}, L.~{Zhao}, L.~{Wu}, L.~{Zhong}, M.~{Liu}, M.~{Huang}, P.~{Zhang}, Q.~{Zheng}, R.~{Lu}, S.~{Duan}, S.~{Zhang}, S.~{Cao}, S.~{Yang}, W.~L. {Tam}, W.~{Zhao}, X.~{Liu}, X.~{Xia}, X.~{Zhang}, X.~{Gu}, X.~{Lv}, X.~{Liu}, X.~{Liu}, X.~{Yang}, X.~{Song}, X.~{Zhang}, Y.~{An}, Y.~{Xu}, Y.~{Niu}, Y.~{Yang}, Y.~{Li}, Y.~{Bai}, Y.~{Dong}, Z.~{Qi}, Z.~{Wang}, Z.~{Yang}, Z.~{Du}, Z.~{Hou}, Z.~{Wang}, {ChatGLM: A Family of Large Language Models from GLM-130B to GLM-4 All Tools}, arXiv e-prints (2024) arXiv:2406.12793\href {http://arxiv.org/abs/2406.12793} {\path{arXiv:2406.12793}}, \href {https://doi.org/10.48550/arXiv.2406.12793} {\path{doi:10.48550/arXiv.2406.12793}}.

\bibitem{ZhuLJZ23}
Z.~Zhu, K.~Lin, A.~K. Jain, J.~Zhou, Transfer learning in deep reinforcement learning: {A} survey, {IEEE} Trans. Pattern Anal. Mach. Intell. 45~(11) (2023) 13344--13362.
\newblock \href {https://doi.org/10.1109/TPAMI.2023.3292075} {\path{doi:10.1109/TPAMI.2023.3292075}}.

\bibitem{Metelli24}
A.~M. Metelli, Recent advancements in inverse reinforcement learning, in: M.~J. Wooldridge, J.~G. Dy, S.~Natarajan (Eds.), Thirty-Eighth {AAAI} Conference on Artificial Intelligence, {AAAI} 2024, Thirty-Sixth Conference on Innovative Applications of Artificial Intelligence, {IAAI} 2024, Fourteenth Symposium on Educational Advances in Artificial Intelligence, {EAAI} 2014, February 20-27, 2024, Vancouver, Canada, {AAAI} Press, 2024, p. 22680.
\newblock \href {https://doi.org/10.1609/AAAI.V38I20.30296} {\path{doi:10.1609/AAAI.V38I20.30296}}.

\end{thebibliography}



\end{document}